# A Lightweight CNN Integrated Compact Convolutional Transformer for Multi-Scale Feature Learning and reducing computational complexity for breast cancer mammography image detection and classification

Md Taimur Ahad **(Corresponding Author)**
Department of Management
North South University, Dhaka, Bangladesh
Email: Taimur.ahad@northsouth.edu

**Ainuddin Ahmed**
Department of Management
North South University, Dhaka, Bangladesh
Email: ainuddin.ahmed.251@northsouth.edu

**A Lightweight CNN Integrated Compact Convolutional Transformer for Multi-Scale Feature Learning and reducing computational complexity for breast cancer mammography image detection and classification**

# Abstract

*This study proposes a novel Convolutional Neural Network integrated Compact Convolutional Transformer (CNN–CCT) model for automated breast cancer classification using mammographic imaging. The model integrates CNNs in CCT to extract both local texture features and global features from mammographic images. The CNN-extracted features are then reshaped into compact patch tokens using a CCT tokenizer, followed by the addition of positional embeddings to preserve spatial structure. A lightweight transformer encoder consisting of multi-head self-attention layers is employed to model long-range dependencies within the token sequence. Finally, a classification head with global average pooling and a dense SoftMax layer produces the final prediction for benign and malignant classes. The model was tested on 3 sets of breast cancer mammography. With only 250,435 parameters, the model consistently achieves strong performance across 5-fold cross-validation experiments on 2-class, 3-class, and 5-class mammographic datasets. The model achieved 99%–100% accuracy across 3 datasets, indicating robust generalization. Furthermore, Explainable AI (XAI) is integrated into the model to explain the breast cancer classification process to enhance clinical trust. The results indicate that the proposed framework is well-suited for computer-aided diagnosis systems, particularly in resource-constrained clinical environments. This study highlights the potential of combining convolutional and transformer-based architectures for improved medical image analysis. The novelty of the proposed CNN-integrated CCT overcomes the limitation of CNN's gradient degradation in the last layers by integrating convolutional tokenization with transformer-based learning. Lighter than ViT, which is effective in capturing long-range dependencies, the model has also proven efficient in breast cancer classification by capturing long-range dependencies among breast tissue regions.*

# 1. Introduction

Breast cancer is one of the foremost global health concerns, as breast cancer significantly contributes to mortality across both industrialized and developing nations. According to the World Health Organization (WHO), approximately one in every ten newly diagnosed cancers worldwide is breast cancer, making it the most frequently diagnosed cancer among women. Recent reports from the International Agency for Research on Cancer (IARC) estimate that about 2.3 million women were diagnosed with breast cancer in 2022, resulting in nearly 670,000 deaths globally. In this context, early and accurate breast cancer detection is critical for a better treatment plan.

Breast cancer often presents subtle abnormalities such as lumps, architectural distortions, or tissue heterogeneity. These symptoms are identified through medical imaging, including mammography. Mammographic images play a vital role in breast cancer screening, specifically in dense breast tissues. The traditional breast cancer detection process involves manually inspecting features such as texture, shape, and intensity. Even mammographic interpretation depends on a doctor or pathologist's manual inspection. While these approaches provided initial improvements, the inspection is often affected by noise, low contrast, and imaging artifacts, which lead to diagnostic variability, human error, and high false-positive rates. Consequently, there is a growing demand for computer-aided diagnosis (CAD) systems that help clinicians improve diagnostic accuracy and consistency.

The advent of deep learning (DL) has significantly enhanced CAD systems by enabling the extraction of features from breast cancer mammography images, thereby improving the precision of cancer detection and classification. Among DL models, Convolutional Neural Networks (CNNs) in particular have demonstrated strong performance in breast cancer classification tasks. However, CNN-based models often struggle to capture long-range contextual dependencies and require substantial amounts of labeled data and computational resources. To fill this gap, transformer-based architectures have emerged as powerful alternatives, thanks to their ability to model global dependencies via self-attention mechanisms.

Among such transformer-based architectures, the Compact Convolutional Transformer (CCT) is a hybrid vision model that combines convolutional neural networks with transformer-based self-attention. It uses a convolutional tokenizer at the input stage to extract local spatial features and convert images into token sequences for transformer encoders (Hassani et al., 2021). This design helps the model learn both local patterns and global dependencies more evenly. CCT performs well in small- to medium-scale data settings where full-scale Vision Transformers often struggle. It also reduces the need for very large datasets by adding useful CNN-style inductive bias. Overall, it is more data-efficient and easier to train than standard Vision Transformers in many practical cases.

However, CCT still has some limitations. Transformer models require large amounts of data and are computationally expensive. The self-attention mechanism remains computationally intensive, especially as input sequences grow longer, leading to higher memory usage and slower inference. Even with convolutional tokenization, it can still be less efficient than lightweight CNN models in resource-constrained environments (Hosseini et al., 2025). The architecture also requires careful tuning of convolutional depth, kernel size, and pooling strategy, making model design less straightforward. Some studies also note that hybrid models may lose flexibility compared to pure transformer designs in certain structured vision tasks (EmergentMind, 2026).

To address these limitations, we propose a lightweight CNN-integrated CCT that efficiently learns both local and global features from mammographic images for breast cancer classification. Breast mammography analysis requires a simultaneous understanding of fine-grained tissue variations and broader anatomical structures. This design is motivated by the nature of breast imaging, in which clinically relevant cues appear at multiple scales, ranging from subtle micro-level patterns, such as microcalcifications, to large structural abnormalities, such as masses and asymmetries. A single DL model might be insufficient to represent both types of features effectively, and therefore, a CNN-integrated CCT is adopted.

In the case of CCT, ablation studies are especially important as the model performance depends on convolutional tokenization, transformer encoder layers, and sequence pooling. Since CCT is designed to balance CNN-style local feature extraction with transformer-based global attention, it is unclear which component is responsible for the gains in accuracy or efficiency. Moreover, this study integrates a lightweight CNN into CCT; therefore, ablation studies help isolate the effects of the framework's components. Without ablation analysis, it is difficult to justify the claim that the hybrid design is truly necessary or optimal, rather than merely empirically effective. In this study, we conducted ablation studies to understand how each component of the CNN integrates CCT to improve performance, efficiency, and generalization in breast cancer detection and classification.

In this study, we have integrated XAI into the lightweight CNN-integrated CCT model. CCT is often treated as a black-box model, as it lacks explainability. This makes it difficult to fully interpret how convolutional tokenization and attention jointly contribute to predictions in CCT. Recent studies on vision transformers also highlight that explanation methods remain limited (Fantozzi & Naldi, 2024; Xie et al., 2023). Existing research on ViTs shows that even attention-based explanations can be unreliable or incomplete for understanding model decisions (Xie et al., 2023).

The contribution of the study is the development of a CNN-integrated CCT framework that combines local feature extraction with global context learning for mammographic breast cancer classification. CNN models are effective in extracting local patterns but may struggle to capture relationships between distant regions. However, ViT is ineffective in identifying small local abnormalities due to early patch-based image division. In mammography, important signs such as microcalcifications, irregular margins, and architectural distortion are often small but related to surrounding tissue structures. Therefore, the proposed model uses convolutional tokenization to preserve meaningful local features before transformer processing. Instead of directly splitting images into fixed patches, the model learns informative feature tokens from convolutional representations, thereby better retaining subtle lesion patterns. The use of stochastic depth and a compact transformer design further improves model stability and reduces overfitting, making the framework suitable for medical image

# 2. Literature Review

Early deep learning approaches relied primarily on standalone CNN architectures for binary or multi-class classification of breast cancer images. For example, Simonyan et al. (2024) demonstrated the effectiveness of CNNs for histopathological classification, achieving competitive accuracy using standard architectures trained on publicly available datasets. Similarly, transfer learning approaches using pre-trained models such as ResNet, VGG, EfficientNet, and MobileNet have shown strong performance in breast cancer detection tasks (Elzaghmouri et al., 2025). These models benefit from learned representations from large-scale datasets such as ImageNet, allowing them to perform well even with limited medical imaging data. However, their performance is often constrained by dataset imbalance and domain shift between natural and medical images.

A major research trend involves enhancing CNNs with hybrid deep learning architectures to improve feature representation and classification accuracy. Kaddes et al. (2025) proposed a CNN–LSTM hybrid model for breast cancer classification, in which the CNN extracts spatial features and the LSTM captures sequential dependencies within the feature maps. The model achieved up to 99.90% accuracy, demonstrating the benefit of integrating temporal modeling with CNN-based spatial learning. Similarly, Brahmareddy and Selvan (2025) introduced TransBreastNet, a CNN–Transformer hybrid model capable of simultaneously classifying breast cancer subtypes and analyzing temporal lesion progression. The integration of transformers allows the model to capture long-range dependencies that CNNs typically struggle with. Sreelekshmi et al. (2024) also proposed a SwinCNN architecture that combines Swin Transformers with CNN layers for improved histopathological grading, achieving strong performance across multiple datasets, including BACH and BreakHis. These studies indicate that hybrid architectures consistently outperform traditional CNN models by combining local feature extraction with global contextual learning.

Another significant direction in the literature is feature fusion using multiple CNN architectures. Chakravarthy et al. (2024) proposed a multi-class classification system using hybrid feature fusion from VGG16, VGG19, ResNet50, and DenseNet121. This approach improves robustness by combining complementary feature representations from different networks. The proposed fusion-based model achieved accuracies exceeding 97% across multiple datasets, including MIAS, CBIS-DDSM, and INbreast. This demonstrates that ensemble CNN approaches can reduce model bias and improve generalization.

Several studies integrate traditional feature extraction techniques with CNN-based models to enhance performance and interpretability. Gül (2025) proposed a hybrid model that combines Local Binary Patterns (LBP) with a CNN for histopathological breast cancer diagnosis. The LBP method enhances texture representation, while CNN provides deep feature learning. Mannarsamy et al. (2025) introduced SIFT-BCD, which integrates the Scale-Invariant Feature Transform (SIFT) with CNN features and fuzzy decision-tree classifiers. The model achieved up to 99.20% accuracy, showing that hybrid classical-deep learning approaches can significantly improve performance in medical image classification. These approaches are particularly useful in histopathological imaging, where texture and structural variations play a critical role in diagnosis.

Recent advancements focus on improving CNN architectures through multi-scale processing, optimization algorithms, and dimensionality reduction techniques. Bohra et al. (2026) proposed a wavelet-CNN fusion architecture that leverages multi-resolution wavelet transforms to extract fine-grained texture features, achieving 99.34% accuracy on the BreakHis dataset. Similarly, Liu et al. (2024) introduced a kernel-based CNN with principal component feature fusion (CNN-PCFF), which reduces dimensionality while preserving discriminative features.

Alzahrani et al. (2025) enhanced CNN performance for thermography-based breast cancer detection by using Particle Swarm Optimization (PSO) to tune hyperparameters, achieving 98.8% accuracy. These studies demonstrate that optimization and multi-scale feature extraction significantly enhance CNN performance in medical imaging tasks.

Recent research has shifted toward integrating CNNs with Transformer architectures to capture both local and global features. Sreelekshmi et al. (2024) proposed SwinCNN, a hybrid model combining Swin Transformers with CNN layers for breast cancer grading. The model achieved strong performance across multiple datasets, demonstrating the advantage of hierarchical attention mechanisms. Abimouloud et al. (2024) introduced Vision Transformer-based CNN models for histopathological classification, demonstrating improved accuracy over standalone CNNs. Similarly, Katayama et al. (2024) highlighted the transition from CNNs to Vision Transformers in breast pathology, emphasizing improved global feature learning. Nayak (2024) proposed RDTNet, a residual deformable attention-based Transformer network achieving up to 99% accuracy across multiple magnification levels. This model demonstrates how attention-based architectures can significantly enhance lesion feature extraction. More advanced models such as BreasTransNeXt (Acikgoz et al., 2026) integrate CNN inductive bias with transformer attention and GAN-based augmentation, achieving high precision and recall (>97%).

These studies confirm that hybrid CNN–Transformer models outperform traditional CNNs in capturing both fine-grained and global contextual features.

Transformers have gained significant attention in medical imaging due to their ability to model long-range dependencies. Goceri (2025) proposed a CNN-Transformer hybrid system with stain normalization for whole-slide image analysis, improving robustness in histopathological classification. Vanitha et al. (2024) introduced attention-based feature fusion using external attention transformers to enhance feature representation in breast cancer images. Abd Elaziz et al. (2024) proposed CrossViT, combined with the Growth Optimizer algorithm, for feature selection in IoMT environments, achieving improved classification accuracy. Furthermore, EAT (External Attention Transformer) models have demonstrated high efficiency, achieving up to 99% accuracy while maintaining low computational complexity. These studies highlight that transformer-based models are particularly effective in capturing global dependencies but often require hybridization to overcome data limitations.

Recent studies have also explored multimodal learning approaches that combine imaging and non-imaging data. Brahmareddy and Selvan (2025) proposed TransBreastNet, a CNN–Transformer hybrid system capable of simultaneously performing subtype classification and lesion progression analysis. The model integrates spatial, temporal, and clinical features, achieving over 95% accuracy. Similarly, research from the MICCAI workshop (2024) demonstrated multimodal fusion of imaging and textual data (radiology reports), showing that late fusion techniques improve classification performance across multiple architectures.

Table 1: Research matrix

| **Author (Year)** | **Method** | **Dataset** | **Accuracy** | **Key Contribution** | **Gap** |
|---|---|---|---|---|---|
| Kaddes et al. (2025) | CNN–LSTM | Kaggle datasets | 99.90% | Spatial + temporal learning | Limited modality diversity |
| Sreelekshmi et al. (2024). | SwinCNN | BACH, BreakHis, IDC | ~98% | Transformer + CNN hybrid | High complexity |
| Chakravarthy et al. (2024). | CNN feature fusion | MIAS, CBIS-DDSM, INbreast | ~98% | Ensemble CNNs | Computational cost |
| Elzaghmouri et al. (2025). | Transfer learning CNNs | Multiple datasets | ~97–98% | Comparative CNN study | Limited innovation beyond baseline |
| Brahmareddy et al. (2025). | CNN–Transformer | Mammogram dataset | 95%+ | Multi-task learning | Needs real-world validation |
| Gül (2025) | LBP + CNN | Histopathology | ~98% | Texture-based enhancement | Limited scalability |
| Alzahrani et al. (2025). | CNN + PSO | Thermography | 98.80% | Optimized CNN tuning | Domain-specific |
| Simonyan et al. (2024). | CNN baseline | Histopathology | ~90–92% | Baseline CNN evaluation | Lower accuracy vs hybrids |
| Fontes et al. (2025). | 3D CNN | MRI dataset | AUC 0.961 | Volumetric analysis | Limited dataset size |
| Bohra et al. (2026). | Wavelet-CNN | BreakHis | 99.34% | Multi-scale features | High computational cost |
| Nayak (2024) | RDTNet (Transformer) | Histopathology | ~99% | Deformable attention transformer | High complexity |
| Sreelekshmi et al. (2024). | SwinCNN | BACH, BreakHis | ~98% | Swin Transformer + CNN | Resource heavy |
| Abimouloud et al. (2024). | ViT-CNN hybrid | BreakHis | ~98% | Hybrid ViT-CNN | Data intensive |
| Feng et al. (2024). | HTBE-Net | Ultrasound | High | Segmentation model | Limited generalization |
| Goceri (2025) | CNN + Transformer | Whole-slide images | ~97–98% | Stain normalization + hybrid model | Complexity |
| Aldawsari et al. (2026). | Swin + Fusion Net | Mammogram | ~98–99% | Multi-scale fusion | High compute cost |
| Vanitha et al. (2024). | Attention Fusion | Histopathology | ~97–98% | External attention transformer | Dataset dependency |
| Katayama et al. (2024). | CNN → ViT review | Pathology | N/A | Review of AI transition | Theoretical |
| Acikgoz et al. (2026). | BreasTransNeXt | Mammogram | ~97% | Hybrid attention system | Needs deployment testing |

| Abd Elaziz et al. (2024). | CrossViT + optimizer | Multi-dataset | High | Feature selection + ViT | Complexity |
|---|---|---|---|---|---|
| Hussain et al. (2024) | Multimodal CNN + ViT | Imaging + text | ~95% | Multimodal fusion | Integration challenges |

# 3. Research methodology

The research methodology (see Figure 1) describes the hardware and software environments, the datasets used in the study, image preprocessing, the CNN-integrated CCT model, and the training process.

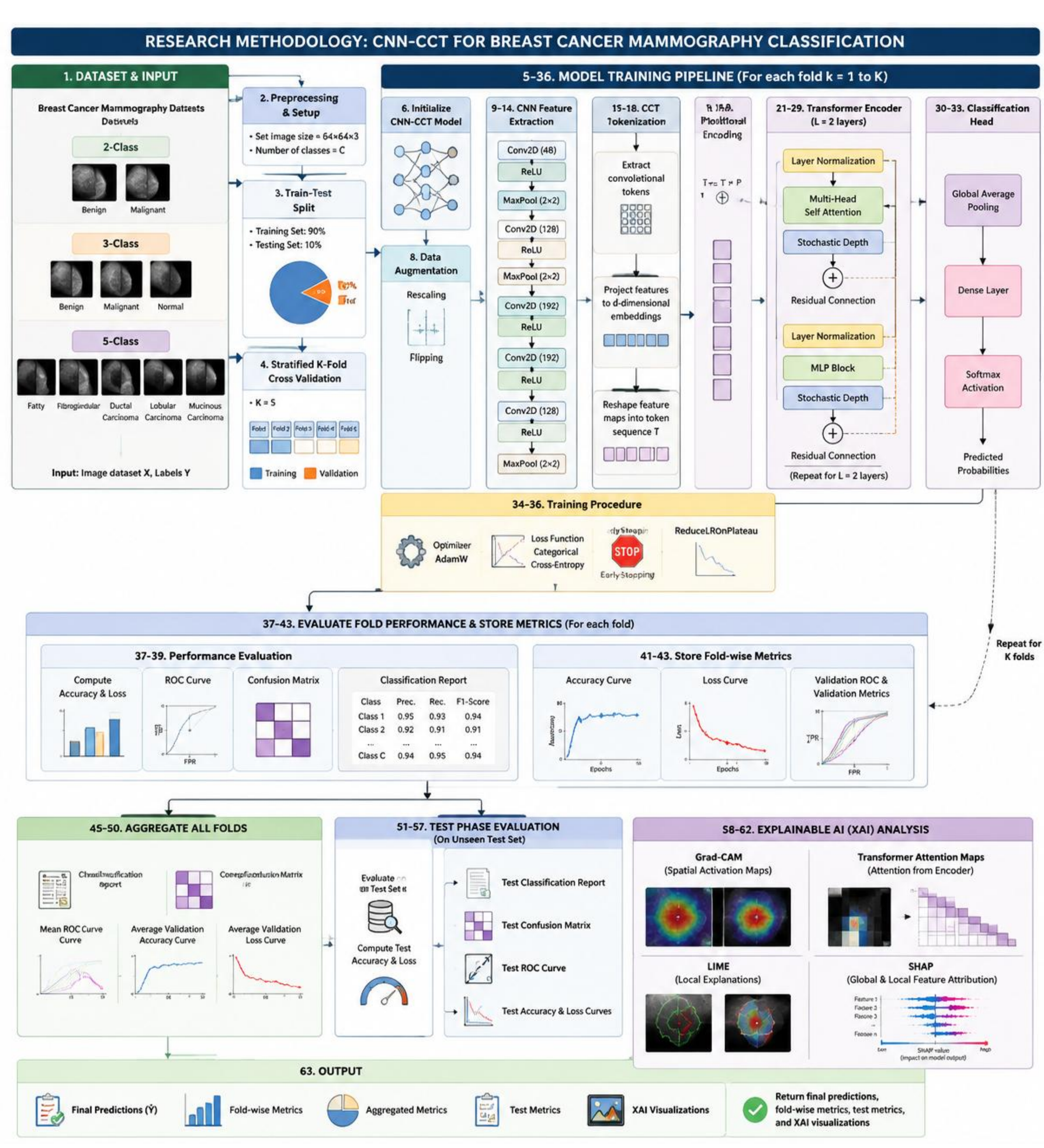

Figure 1: Research methodology adopted in this research

## 3.1 Dataset description

To ensure performance consistency and generalizability, the CNN-Integrated CCT model was trained and tested on both balanced and imbalanced datasets with varying numbers of breast cancer modalities (See Figure 2). Furthermore, it was essential to analyze how the model performs in detecting and classifying unseen breast cancer images after training on the training data. These strategies will provide a better understanding of the CNN-Integrated CCT model construction strategy, especially across classes that are not evenly distributed.

Dataset A consists of 7,632 mammogram images categorized into two classes: 2,520 benign and 5,112 malignant (Huang & Lin, 2020). The mammography images were originally collected from the Breast Center at the Centro Hospitalar de S. Joao (CHSJ) in Porto. Dataset B consists of 3 classes: Malign, Normal, and Benign. The link to the dataset is https://www.kaggle.com/datasets/emiliovenegas1/mammography-dataset-from-inbreast-mias-and-ddsm

Dataset C is MammoNet32k, a comprehensive, standardized mammography dataset comprising 32,191 high-quality mammographic images from 7,079 unique patients across 7 major public datasets. All images converted to 16-bit grayscale PNG at 1024×1024 resolution. The dataset combines data from the US, Europe, China, and the Middle East. Five classes were included: CBIS, DSM-5, DMID, Inbreast, and Kau BCDM. The source of the dataset is https://www.kaggle.com/datasets/theosmithdevey/mammonet20k.

In this study, 70% of the images were used to train the CNN-Integrated CCT, 20% for validation, and 10% for testing on unseen images. To prevent data leakage, the photos were split into train, validation, and test directories. As the authors [53] suggested, the CNN-Integrated CCT model verification stage should include evaluating the final model on a dataset unseen during training. The study also indicated that 'test set' and 'validation set' should not be considered the same.

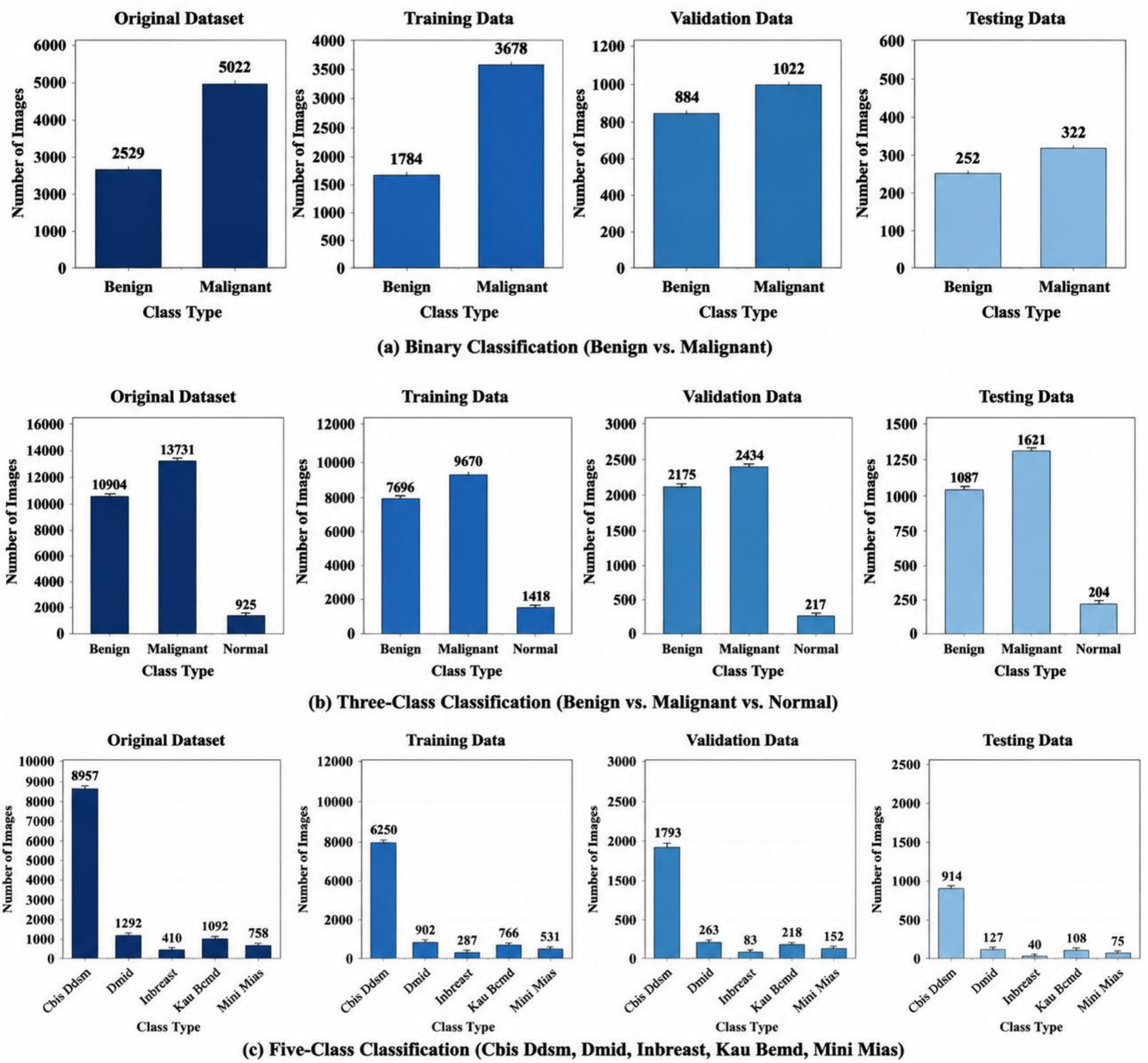

Figure 2: Data distribution in 3 datasets

## 3.2 Hardware Specification

The experiments were conducted via the Precision 7680 Workstation. The workstation is a 13th-generation Intel® Core™ i9-13950HX vPro with Windows 11 Pro, an NVIDIA® RTX™ 3500 Ada Generation GPU, 32 GB DDR5 RAM, and a 1 TB SSD. Python (version 3.9) was chosen as the programming language because it supports TensorFlow-GPU, SHAP, and LIME generation.

## 3.3 Image preprocessing

To enhance the visibility of imaging features and reduce noise, a multistep preprocessing pipeline was applied. Contrast Limited Adaptive Histogram Equalization (CLAHE) was employed to improve local contrast in dense tissue regions, followed by Gaussian blurring to suppress high-frequency noise, defined as:

$$G(x,y) = 1\ 2\pi\sigma 2 \exp -x2 + y2\ 2\sigma 2\ (1)$$

Where σ represents the standard deviation of the Gaussian distribution, subsequently, bilateral filtering was applied to preserve edge structures critical for mass detection, expressed as:

$$Ifiltered(x) = 1\ Wp \sum xi \in \Omega\ I(xi) \cdot fr(|I(xi) - I(x)|) \cdot fs(|xi - x|)(2)$$

where fr denotes the range kernel, fs the spatial kernel, and Wp the normalization factor. Non-local means denoising was used to enhance smoothness further while preserving important features. Finally, unsharp masking was performed to improve edge clarity, formulated as:

$$Isharp = Ioriginal + \alpha \cdot (Ioriginal - Gblur)$$

with $\alpha$ =1.5 controlling sharpness. (3)

## 3.4 Image augmentation

In this study, we employed extensive data augmentation techniques, including rotation, flipping, and contrast modulation, to generate a richer and more diverse dataset for Mammography-based breast cancer classification (see Figure 3). Image augmentation has proven to be a highly effective and practical approach for enhancing the robustness of models with limited ground-truth datasets [54]. In this study, we applied rotations, scaling, flipping, contrast adjustments, and noise addition on the training datasets. In the context of breast cancer detection, these techniques enable models to better account for variability in cancer characteristics, including size, shape, and location. For example, rotation (±20°) allows the model to handle cancers at different orientations, while scaling (±10%) enables it to handle variations in cancer size. Additionally, random flipping helps the model generalize across different image orientations, ensuring it can identify cancers in both left- and right-hemisphere scans. Contrast adjustment (±15%) enables the model to adapt to differences in image quality, while Gaussian noise (variance of 0.05) improves robustness against imaging artifacts often found in medical scans.

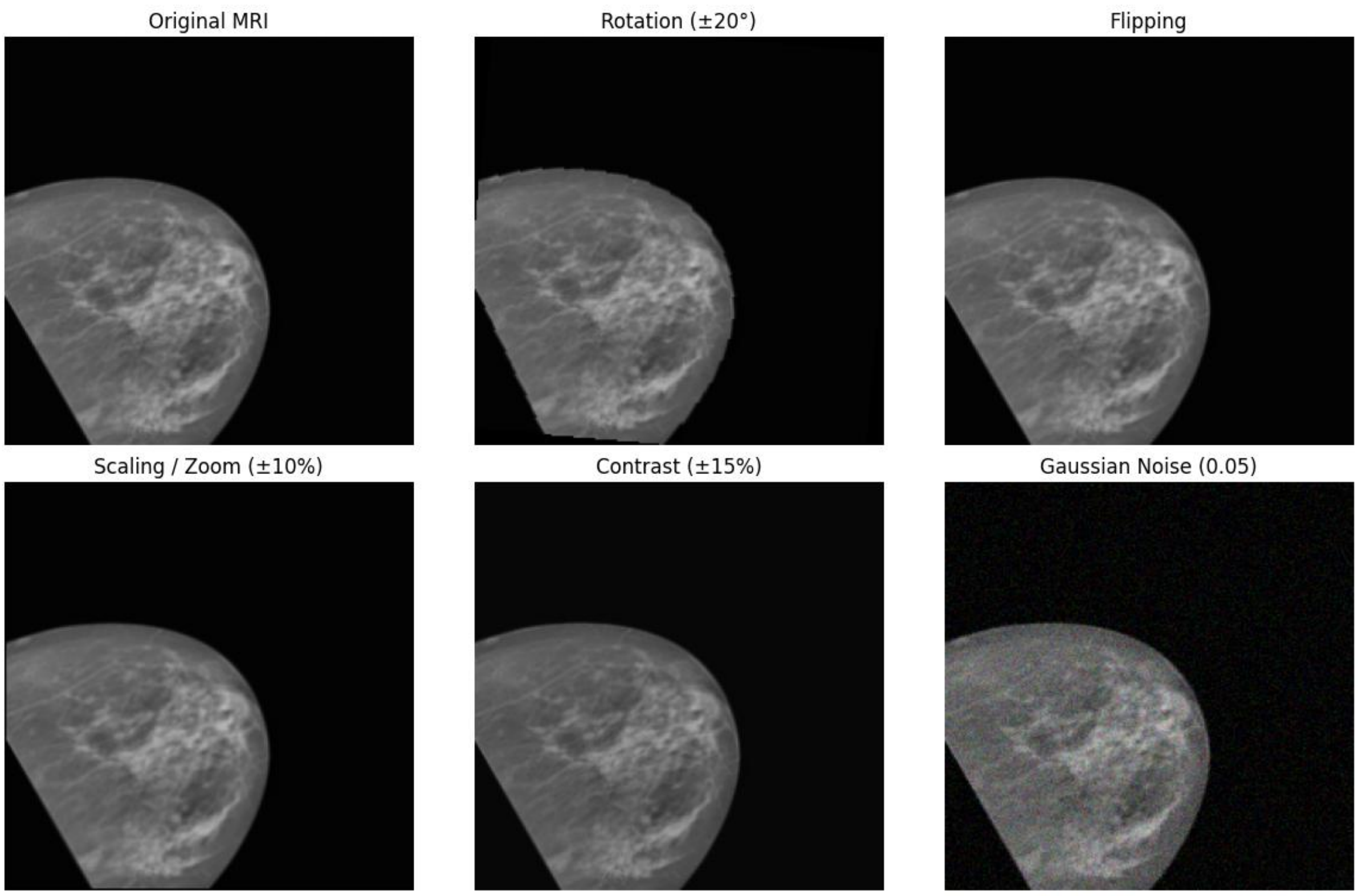

Figure 3. Samples of augmentation

## 3.5 Description of proposed CNN-Integrated CCT Model

The proposed model combines convolutional feature extraction with transformer-based global attention. The CNN-integrated CCT model can be expressed as a sequential transformation: the mammogram image is first encoded by a convolutional feature extractor, then transformed into token representations, followed by global contextual modeling with a transformer encoder, and finally mapped to diagnostic classes via a classification head. The CNN block extracts local breast tissue patterns. The CCT block converts these patterns into tokens and learns global relationships between mammographic regions. The complete model flow is:

$$Input \rightarrow DataAugmentation \rightarrow CNNFeatureExtractor \rightarrow CCTTokenizer$$
$$\rightarrow PositionalEmbedding \rightarrow TransformerEncoder$$
$$\rightarrow GlobalAveragePooling \rightarrow SoftmaxClassifier$$

A more complete expression is:

$$\hat{y} = Softmax\left(Dense\left(GAP\left(Transformer\left(Tokenizer\left(CNN\big(A(X)\big)\right) + P\right)\right)\right)\right)$$

where **X** is the input mammogram image. $A(X)$is the augmented image. $CNN(.)$extracts local feature maps. $Tokenizer(.)$converts the feature maps into token embeddings. $Transformer(.)$learns global contextual relationships. $Softmax(.)$produces the final diagnostic probability.

**Input Layer**

The input mammogram image can be represented as $X \in R^{(H\times W\times 3)}$. Here, H is the image height, W is the image width, and 3 represents the RGB channels. Although mammograms are usually grayscale, they are converted to RGB format to match the model's input requirements. This provides a uniform input format for training.

**Data Augmentation Layer**

This layer is important for mammography classification because breast position, compression, and acquisition angle may vary across patients. The augmentation process can be expressed as $X_{aug} = A(X)$. In the implementation, $X' = X/255$ and $X_{aug} = RandomHorizontalFlip(X')$. Here $X'$ is the rescaled image and $X_{aug}$ is the augmented image. Rescaling pixel values from 0–255 to 0–1. This improves numerical stability during training.

**CNN Feature Extraction Block**

After data augmentation, the mammogram is passed through the CNN feature extractor. This CNN block is placed before the CCT tokenizer. It extracts local and fine-grained features from the mammogram. The CNN block follows this structure:

$$Conv2D(48) \rightarrow MaxPooling \rightarrow Conv2D(128) \rightarrow MaxPooling \rightarrow Conv2D(192) \\ \rightarrow Conv2D(192) \rightarrow Conv2D(128) \rightarrow MaxPooling$$

The CNN feature extraction process can be written as $F_{\mathrm{CNN}} = CNN\left(X_{aug}\right)$. Here $F_{CNN}$ is the final convolutional feature map. A convolutional layer can be expressed as:

$$F_l = ReLU\left(W_l * F_{(l-1)} + b_l\right)$$

where $F_l$is the output feature map of layer **l**. $W_l$ is the convolution kernel. $b_l$ is the bias term. The symbol * represents convolution. **ReLU** is the activation function.

The first convolutional layer uses 48 filters. It captures low-level features such as edges, curves, small texture changes, and intensity variations. The second convolutional layer uses 128 filters. It learns more detailed tissue patterns, including density variations and abnormal local textures. The next two convolutional layers use 192 filters. These layers learn deeper and more complex features. They help detect irregular lesion margins, mass boundaries, and structural distortion. The final convolutional layer uses 128 filters. It refines the feature representation before passing it to the CCT tokenizer. Max pooling is applied after selected convolutional layers. It can be written as:

$$F_{pool} = MaxPool(F_l)$$

Max pooling reduces spatial size while preserving the strongest responses. It also reduces computational cost. This is useful in mammography because suspicious patterns can appear in slightly different positions. The CNN block is crucial for breast cancer screening mammography. Many cancer-related signs are small and local. These include microcalcification-like bright spots, tissue density changes, mass edges, spiculated margins, and local architectural distortion. The CNN block helps the model capture these subtle visual features before transformer processing.

**CCT Tokenizer**

After CNN feature extraction, the feature map is passed into the CCT tokenizer. The tokenizer converts the CNN feature map into a sequence of tokens. The tokenizer operation can be written as:

$$F_{tok} = Pool\left(ReLU\left(Conv(F_{CNN})\right)\right)$$

where $F_{tok}$ is the tokenized convolutional feature map. In the model, the tokenizer uses convolutional layers followed by max-pooling. This is different from a standard Vision Transformer, which directly divides an image into fixed patches. The convolutional tokenizer is better suited to mammography because it preserves local spatial information before tokenization. This is important because mammographic abnormalities are often small and subtle. If the image is divided into large fixed patches too early, fine details may be weakened. The convolutional tokenizer helps retain local lesion-related features.

**Projection to Token Embedding Space**

The tokenizer's output is projected into a fixed embedding dimension using a dense layer. This step prepares the features for transformer processing. The projection can be written as:

$$E = Dense(F_tok)$$

where **E** is the projected feature representation, then the projected feature map is reshaped into a sequence:

$T = Reshape(E)$ or, $T = [t_1, t_2, t_3, \dots, t_N]$where **T** is the token sequence. $t_i$is one token. **N** is the total number of tokens. Each token represents a local mammographic region after CNN and tokenizer-based feature extraction.

**Positional Embedding**

The transformer does not naturally understand token positions. Therefore, positional embedding is added to the token sequence. This can be written as:

$Z_0 = T + P$ where $Z_0$ is the position-aware token sequence. **T** is the token embedding. **P** is the learnable positional embedding. For each token:

$z_i = t_i + p_i$where $t_i$ is the token embedding and $p_i$is its positional embedding.

This step is important in mammography because the location of abnormal tissue matters. Breast cancer diagnosis depends not only on texture but also on lesion position, tissue distortion, and relationships among surrounding regions.

**Transformer Encoder Block**

The position-aware tokens are passed through transformer encoder layers. Each transformer block contains layer normalization, multi-head self-attention, residual connection, stochastic depth, and an MLP block.

The input to a transformer layer can be written as:

$Z_l = TransformerLayer(Z_{(l-1)})$where $Z_l$ is the output of the transformer layer **l**.

Layer normalization is first applied:

$Z_norm = LayerNorm(Z_{(l-1)})$Layer normalization stabilizes training and improves convergence.

**Multi-Head Self-Attention**

The transformer uses multi-head self-attention to learn relationships between different mammographic regions. The attention mechanism is written as:

$Attention(Q, K, V) = Softmax(QK^T/\sqrt{d_k})V$where **Q** is the query matrix, and K is the key matrix. **V** is the value matrix. $d_k$ is the key dimension. Multi-head self-attention can be written as:

$MHSA(Z) = MultiHeadAttention(Z, Z)$This allows each token to attend to all other tokens. It helps the model learn global relationships across the mammogram.

This is useful because malignant signs may not be completely local. A suspicious region may need to be interpreted together with surrounding tissue density, asymmetry, and architectural distortion. Self-attention allows the model to capture these long-range dependencies.

**Residual Connection and Stochastic Depth**

After self-attention, the output is added back to the input using a residual connection. Stochastic depth is applied for regularization. This can be written as:

$$Z_att = Z_{(l-1)} + SD\left(MHSA\left(LayerNorm(Z_{(l-1)})\right)\right)$$

where **SD(.)** represents stochastic depth. Stochastic depth randomly drops residual branches during training. It can be written as:

$$SD(x) = x/p, withprobabilityp$$

$$SD(x) = 0, withprobability1 - p$$

where **p** is the survival probability. In the model, the stochastic depth rate increases gradually across transformer layers. This improves regularization and reduces overfitting. It is useful for mammography datasets because medical image datasets are often limited in size.

**MLP Block with GELU Activation**

After attention, the model applies another layer normalization followed by an MLP block. This layer

can be written as $Z_m lp = MLP(LayerNorm(Z_a tt))$. The MLP operation is:

$$MLP(x) = Dense\left(GELU(Dense(x))\right).$$

In the model, the MLP hidden units are:

$$[2 \times projection_d im, projection_d im]$$

The final output of the transformer block is:

$$Z_l = Z_a tt + SD\left(MLP(LayerNorm(Z_a tt))\right)$$

The MLP block learns nonlinear relationships among token features. This helps the model distinguish between benign and malignant tissue patterns that may appear visually similar.

**Final Layer Normalization**

After all transformer layers, layer normalization is applied again. The layer normalization is expressed as $Z_f inal = LayerNorm(Z_L)$. here $Z_L$ is the output of the last transformer layer. This produces a stable final token representation before pooling.

**Global Average Pooling**

The model applies global average pooling over all tokens. Global average pooling allows all token features to contribute to the final prediction. The model does not use a separate class token. This makes the architecture simpler and compact. If **g** is the image-level feature vector, $N$is the number of tokens, $z_i$is the representation of a token $i$, $g = GAP(Z_f inal)$ or $g = (1/N)\Sigma z_i$ .

### Softmax Classification Layer

The pooled feature vector is passed into a dense classification layer. The final prediction is produced using softmax activation. The layer can be written as $\hat{y} = Softmax(W_c g + b_c)$where $\hat{\mathbf{y}}$ is the predicted probability distribution. $W_c$is the classifier weight matrix. $b_c$ is the classifier bias. The predicted $Class = argmax(\hat{y})$. For binary classification, the classes may be benign and malignant. For multi-class classification, the model predicts the probability of each diagnostic category.

### Loss Function

The model is trained using categorical cross-entropy loss. This loss function measures the difference between the true label and the predicted probability distribution. The loss function can be expressed as $L = -\Sigma y_i log(\hat{y}_i)$. here $y_i$is the true class label and $\hat{y}_i$is the predicted probability for the class $i$.

### Optimizer

AdamW, with adaptive learning and weight decay regularization, was used as the optimizer in the model. The update rule can be written as:

$\theta_{(t+1)} = \theta_t - \eta m_t / (\sqrt{v_t} + \varepsilon) - \lambda \theta_t$where $\theta_t$ represents model parameters. $\boldsymbol{\eta}$ is the learning rate. $m_t$ is the first moment estimate. $v_t$ This is the second moment estimate. $\boldsymbol{\lambda}$ is the weight decay coefficient. Weight decay helps reduce overfitting. Gradient clipping is also used to improve training stability.

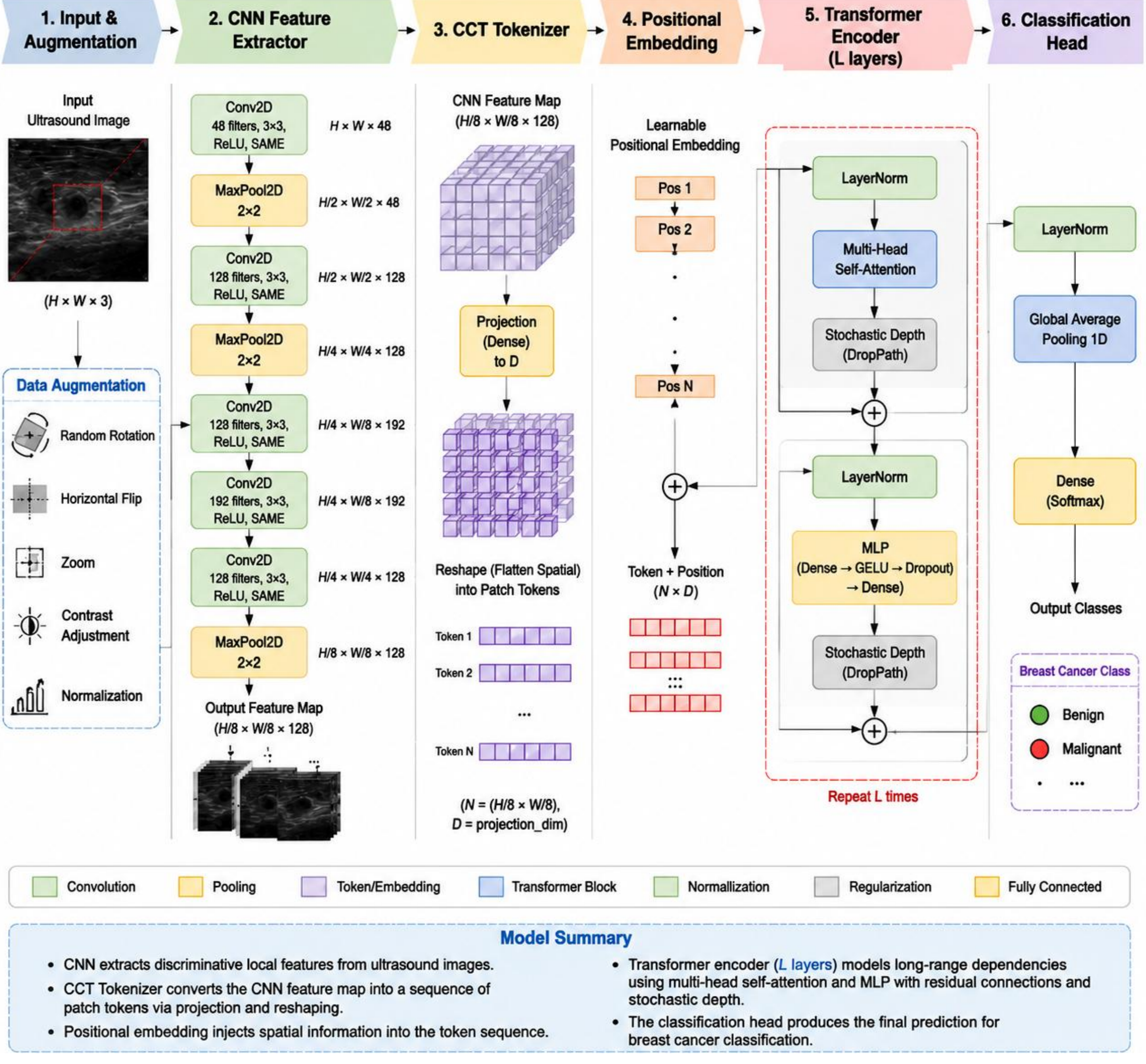

Figure 4: Layer-wise visualization of the CNN-integrated CCT

## The algorithm of the proposed

Input: Image dataset X, Labels Y
Output: Final predicted labels Ŷ and evaluation metrics

1: Initialize dataset X and Y
2: Set image size = 64×64×3, number of classes = C
3: Split the dataset into training (90%) and testing (10%)
4: Apply Stratified K-Fold Cross Validation (K = 5)

5: for each fold k = 1 to K do
6: Initialize CNN-CCT model
7:
8: Apply data augmentation (rescaling, flipping)
9: CNN Feature Extraction:
10: Conv2D(48) → ReLU → MaxPool(2×2)
11: Conv2D(128) → ReLU → MaxPool(2×2)
12: Conv2D(192) → ReLU
13: Conv2D(192) → ReLU
14: Conv2D(128) → ReLU → MaxPool(2×2)
15: CCT Tokenization:
16: Extract convolutional tokens
17: Project features to d-dimensional embeddings
18: Reshape feature maps into a token sequence T
19: Positional Encoding:
20: T ← T + P
21: Transformer Encoder (L = 2 layers):
22: Apply Layer Normalization
23: Multi-Head Self Attention
24: Apply Stochastic Depth
25: Residual Connection
26: Apply Layer Normalization
27: MLP Block
28: Apply Stochastic Depth
29: Residual Connection
30: Global Representation:
31: Apply Global Average Pooling
32: Classification:
33: Dense layer + Softmax activation
34: Train model using AdamW optimizer
35: Optimize using categorical cross-entropy loss
36: Apply EarlyStopping and ReduceLROnPlateau
37: Evaluate fold performance:
38: Compute accuracy, loss, ROC curve, Generate confusion matrix, Generate classification report
41: Store fold-wise metrics:
42: Accuracy curve, loss curve, Validation ROC, validation metrics
44: end for
45: Aggregate all folds:
46: Compute average classification report, averaged confusion matrix, mean ROC curve, average validation accuracy curve, validation loss curve
51: Test Phase Evaluation:
52: Evaluate model on unseen test set, Compute test accuracy and loss, test classification report, test confusion matrix
56: Generate test ROC curve, Compute test accuracy and loss curves
58: Explainable AI Analysis:
59: Apply Grad-CAM for spatial activation mapping
60: Extract attention maps from the transformer encoder
61: Apply LIME for local explanations
62: Apply SHAP for global and local feature attribution
63: Return final predictions, fold-wise metrics, test metrics, and XAI visualizations
END

## 3.6 Training Process of CNN-Integrated CCT Model

The training process for the proposed CNN-Integrated CCT begins with the mammography dataset, which is organized into class-wise directories. Each image is assigned a categorical label based on its folder index, where the label belongs to a finite set of diagnostic classes defined as: $y_i \in 0,1,\ldots,C-1$

Here, C represents the total number of breast cancer classes in the dataset. The images were converted to RGB format and resized to a fixed resolution to ensure uniform input representation across the model. This preprocessing step can be expressed as $I \in R^{(64\times64\times3)}$.

To ensure numerical stability during training and reduce variability across different imaging devices, pixel intensities are normalized to the range [0, 1]. This normalization is defined as: $I' = I/255$

After preprocessing, the dataset is split into training, validation, and testing subsets. A stratified sampling strategy is applied to preserve the original class distribution across all subsets. This ensures that the probability distribution of classes remains approximately consistent, which can be expressed as: $P_train(y) \approx P_test(y)$

To improve the robustness and generalization ability of the model, K-fold cross-validation is applied on the training data. In this setting, the dataset is divided into K equal folds, typically K = 5. For each iteration, the model is trained on K − 1 folds and validated on the remaining fold, which is mathematically represented as:

$$D_train^{(k)} \cup D_val^{(k)} = D_train$$

This strategy ensures stable performance estimates, reduces bias due to data imbalance, and improves generalization to unseen samples.

To further enhance generalization and reduce overfitting, data augmentation is applied in real time during training. Each input image undergoes stochastic transformations such as horizontal flipping and intensity rescaling. This augmentation process is defined as:

$$I_aug = A(I)$$

where A(·) denotes the augmentation function that simulates variations in breast positioning, imaging angle, and scanner noise, thereby improving model robustness to real-world variations.

The training pipeline follows a hierarchical learning strategy beginning with convolutional feature extraction. In this stage, the augmented input image is passed through a convolutional network to extract local visual patterns such as edges, textures, and micro-calcifications. This process is defined as:

$$F_C NN = f_c onv(I_a ug)$$

The extracted convolutional feature maps are then converted into a sequential representation using the CCT tokenizer. This transformation flattens and projects spatial features into a token sequence suitable for transformer processing. It is expressed as:

$$T = Flatten\left(W_p F_C NN\right)$$

where $W_p$ is a learnable projection matrix that maps convolutional features into token embeddings. The resulting token sequence is then enhanced with positional information and passed through a transformer encoder, which learns global dependencies across breast regions. This stage is represented as:

$$Z = Transformer(T + P)$$

where P denotes positional embeddings that preserve spatial awareness, enabling the model to capture anatomical structure, symmetry between breasts, and long-range dependencies across tissue regions.

After global feature modeling, the transformer output is aggregated using global average pooling, and the final classification is performed using a softmax classifier. The prediction function is defined as:

$$\hat{y} = Softmax\left(W \cdot GAP(Z)\right)$$

The training process is optimized with the AdamW optimizer, which updates model parameters via adaptive moment estimation with weight decay. The update rule is given by:

$$\theta_{t+1} = \theta_t - \eta \cdot \left(m_t/\left(\sqrt{v_t} + \varepsilon\right)\right) - \lambda\theta_t$$

where η is the learning rate, λ is the weight decay coefficient, and m_t, v_t are the first and second moment estimates, respectively. The model is trained using categorical cross-entropy loss, which measures the divergence between predicted probabilities and true class labels. The loss function is defined as:

$$L = -\sum_{i=1}^{C} y_i log(\hat{y}_i)$$

During training, performance is continuously monitored on the validation set, and the model achieving the lowest validation loss is selected as the best checkpoint. This selection criterion is defined as:

$$BestModel = argmin(L_val)$$

This complete training strategy ensures that the model learns robust, generalizable, and clinically meaningful representations for mammography classification by effectively combining local convolutional feature extraction with global transformer-based reasoning. Final Training Hyperparameters

Table 2: Hyperparameters of the CNN-integrated CCT model

| **Parameter** | **Value** |
|---|---|
| Image Size | 64 × 64 × 3 |
| Epochs | 250 |
| Batch Size | 32 |
| Learning Rate | $1 \times 10^{-4}$ |
| Weight Decay | $1 \times 10^{-4}$ |
| Optimizer | AdamW (fallback: Adam) |
| Loss Function | Categorical Cross-Entropy |
| K-Folds | 5 |
| CNN Conv Layers | 2 (base configuration) |
| Transformer Layers | 2 |
| Projection Dimension | 64 |
| Attention Heads | 4 |
| Stochastic Depth Rate | 0.1 |
| Positional Embedding | Enabled |

The proposed CCT model is effective for mammography-based breast cancer classification because it combines three useful properties:

First, the convolutional tokenizer captures local mammographic features. These local patterns are important because suspicious findings may appear as small bright microcalcification clusters, irregular mass margins, or localized density changes. The convolution operation learns these local patterns through small receptive fields.

Second, the transformer encoder captures global contextual relationships. This is important because mammography interpretation often depends on the relationship between a suspicious region and the surrounding breast structure. For example, architectural distortion is a global pattern of tissue deformation rather than only a single local texture. Self-attention allows each token to compare itself with all other regions. $z_i \rightarrow \{z_1, z_2, \dots, z_N\}$. Therefore, the model learns both.$local\ lesion - level\ information \qquad global\ breast - level\ context$

Third, stochastic depth and dropout improve generalization. This is important in medical imaging because annotated datasets are often limited, and class imbalance is common. Regularization helps reduce overfitting and improves robustness when the model is tested on unseen mammograms.

The proposed model employs a Compact Convolutional Transformer architecture for breast cancer mammography classification. Given an input mammogram, data augmentation is first applied to improve robustness against image-level variations. The augmented image is then processed by a convolutional tokenizer consisting of repeated convolution, ReLU activation, and max-pooling operations. This tokenizer extracts local mammographic features and converts the image into a compact sequence of visual tokens. Compared with direct patch tokenization, convolutional tokenization introduces an inductive bias that preserves local spatial information, which is important for detecting subtle mammographic structures such as microcalcifications, irregular lesion boundaries, and local density variations.

The resulting token sequence is projected into a fixed-dimensional embedding space and combined with learnable positional embeddings. The encoded sequence is then processed using multiple transformer encoder blocks. Each encoder block contains layer normalization, multi-head self-attention, residual connections, stochastic depth, and an MLP module with GELU activation. The self-attention mechanism models long-range dependencies among mammographic regions, allowing the network to learn both local lesion-specific patterns and global breast-structure relationships. After the final transformer layer, global average pooling aggregates the token representations into a single image-level descriptor. Finally, a dense softmax classifier estimates the probability distribution over the target breast cancer classes.

# 4. Result of the proposed model

The performance of the models is evaluated using the following metrics:

$$Accuracy = ((TP + TN))/((TP + FP + FN + TN))$$

$$Precision = TP/(TP + FP)$$

$$Recall = TP/(TP + FN)$$

$$F1\ score = 2 \times (Precision \times Recall)/(Precision + Recall)$$

$$Specificity = TN/(TN + FP)$$

Furthermore, the accuracy curve, data loss curve, confusion matrix, and XAI were used to visualize the models' performance.

## 4.1 Performance of 5-folds of the CNN-Integrated CCT Model

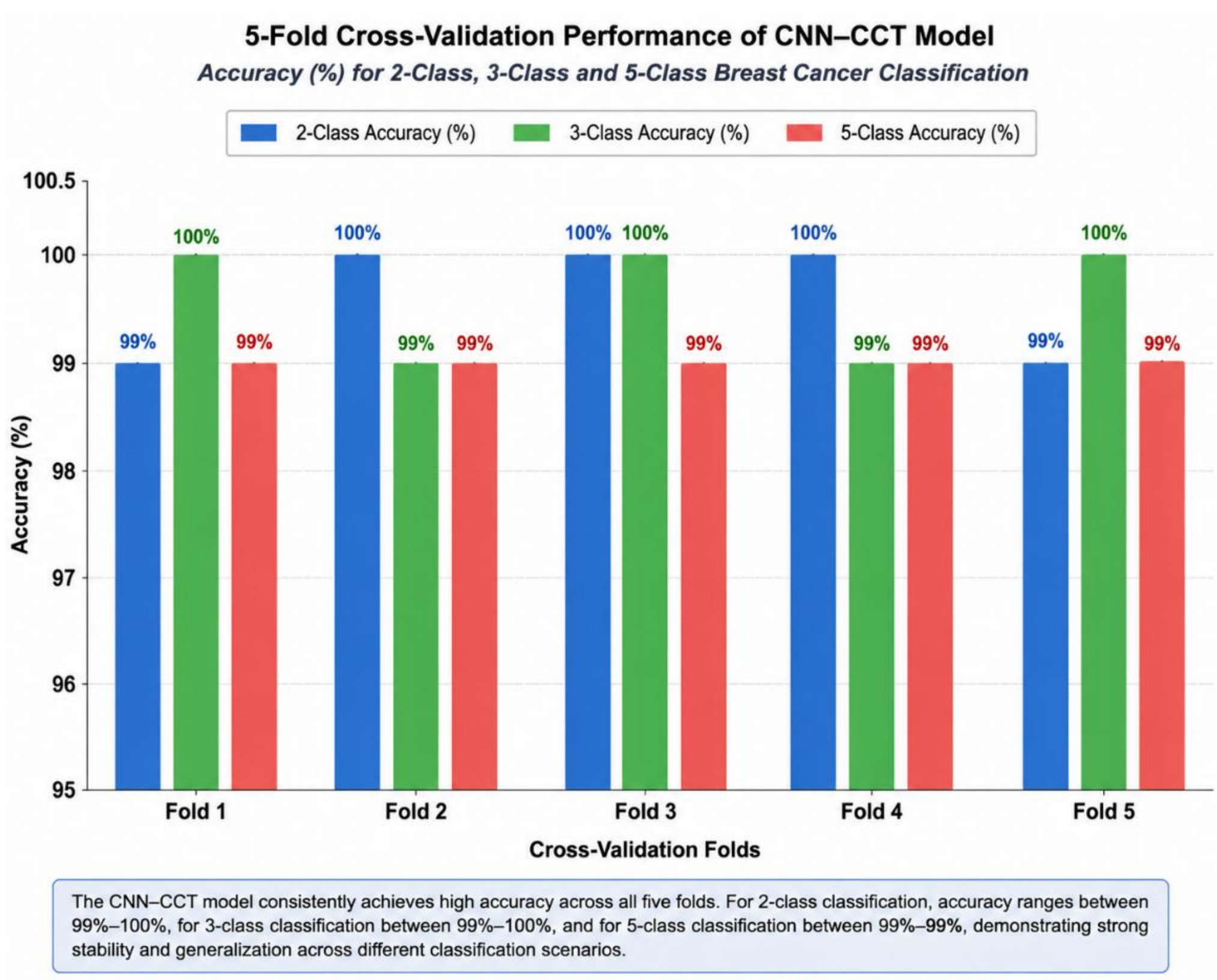

Figure 5: 5 Fold Cross-Validation Performance of CNN-integrated CTT Model

Table 3: Classification report of the 5-fold of the model

| Fold | Class name | Precision | Recall | F1-score | Support | Class name | Precision | Recall | F1-score | Support | Class name | Precision | Recall | F1-score | Support |
|---|---|---|---|---|---|---|---|---|---|---|---|---|---|---|---|
| 1 | Benign | 100% | 99% | 99% | 454 | Benign | 99% | 99% | 99% | 1956 | CBIS DSM | 99% | 99% | 99% | 514 |
| | Malignant | 99% | 100% | 100% | 920 | Malignant | 100% | 100% | 100% | 2468 | Dmid | 100% | 100% | 100% | 183 |
| | | | | | | Normal | 100% | 100% | 100% | 365 | Inbreast | 98% | 97% | 98% | 148 |
| | | | | | | | | | | | Kau Bcmd | 99% | 99% | 99% | 429 |
| | | | | | | | | | | | Mini Mias | 96% | 95% | 96% | 58 |
| | Accuracy | | | 99% | 1374 | Accuracy | | | 100% | 4789 | Accuracy | | | 99% | 1332 |
| | Macro Avg | 99% | 99% | 99% | 1374 | Macro Avg | 100% | 100% | 100% | 4789 | Macro Avg | 99% | 98% | 98% | 1332 |
| | Weighted Avg | 99% | 99% | 99% | 1374 | Weighted Avg | 100% | 100% | 100% | 4789 | Weighted Avg | 99% | 99% | 99% | 1332 |
| 2 | Benign | 100% | 99% | 99% | 454 | Benign | 99% | 99% | 99% | 1956 | CBIS DSM | 99% | 99% | 99% | 514 |
| | Malignant | 99% | 100% | 100% | 920 | Malignant | 99% | 100% | 99% | 2468 | Dmid | 99% | 100% | 100% | 184 |
| | | | | | | Normal | 100% | 99% | 100% | 364 | Inbreast | 99% | 99% | 99% | 147 |
| | | | | | | | | | | | Kau Bcmd | 100% | 99% | 99% | 429 |
| | | | | | | | | | | | Mini Mias | 97% | 97% | 97% | 58 |
| | Accuracy | | | 100% | 1374 | Accuracy | | | 99% | 4788 | Accuracy | | | 99% | 1332 |
| | Macro Avg | 100% | 99% | 100% | 1374 | Macro Avg | 100% | 99% | 99% | 4788 | Macro Avg | 99% | 99% | 99% | 1332 |
| | Weighted Avg | 100% | 100% | 100% | 1374 | Weighted Avg | 99% | 99% | 99% | 4788 | Weighted Avg | 99% | 99% | 99% | 1332 |
| 3 | Benign | 100% | 100% | 100% | 454 | Benign | 100% | 100% | 100% | 1956 | CBIS DSM | 99% | 100% | 99% | 515 |
| | Malignant | 100% | 100% | 100% | 920 | Malignant | 100% | 100% | 100% | 2468 | Dmid | 99% | 100% | 100% | 184 |
| | | | | | | Normal | 100% | 100% | 100% | 364 | Inbreast | 97% | 98% | 97% | 147 |
| | | | | | | | | | | | Kau Bcmd | 99% | 97% | 98% | 428 |
| | | | | | | | | | | | Mini Mias | 98% | 100% | 99% | 58 |
| | Accuracy | | | 100% | 1374 | Accuracy | | | 100% | 4788 | Accuracy | | | 99% | 1332 |
| | Macro Avg | 100% | 100% | 100% | 1374 | Macro Avg | 100% | 100% | 100% | 4788 | Macro Avg | 99% | 99% | 99% | 1332 |
| | Weighted Avg | 100% | 100% | 100% | 1374 | Weighted Avg | 100% | 100% | 100% | 4788 | Weighted Avg | 99% | 99% | 99% | 1332 |
| 4 | Benign | 100% | 99% | 100% | 453 | Benign | 99% | 100% | 99% | 1956 | CBIS DSM | 99% | 100% | 99% | 514 |
| | Malignant | 100% | 100% | 100% | 920 | Malignant | 100% | 99% | 99% | 2467 | Dmid | 100% | 100% | 100% | 184 |
| | | | | | | Normal | 100% | 99% | 99% | 365 | Inbreast | 99% | 99% | 99% | 148 |
| | | | | | | | | | | | Kau Bcmd | 100% | 99% | 99% | 428 |
| | | | | | | | | | | | Mini Mias | 98% | 95% | 96% | 58 |
| | Accuracy | | | 100% | 1373 | Accuracy | | | 99% | 4788 | Accuracy | | | 99% | 1332 |
| | Macro Avg | 100% | 100% | 100% | 1373 | Macro Avg | 100% | 99% | 99% | 4788 | Macro Avg | 99% | 99% | 99% | 1332 |
| | Weighted Avg | 100% | 100% | 100% | 1373 | Weighted Avg | 99% | 99% | 99% | 4788 | Weighted Avg | 99% | 99% | 99% | 1332 |
| 5 | Benign | 100% | 99% | 99% | 453 | Benign | 99% | 100% | 100% | 1955 | CBIS DSM | 100% | 100% | 100% | 514 |
| | Malignant | 99% | 100% | 100% | 920 | Malignant | 100% | 100% | 100% | 2468 | Dmid | 100% | 100% | 100% | 184 |
| | | | | | | Normal | 100% | 99% | 100% | 365 | Inbreast | 99% | 96% | 98% | 148 |
| | | | | | | | | | | | Kau Bcmd | 99% | 99% | 99% | 428 |
| | | | | | | | | | | | Mini Mias | 93% | 98% | 96% | 58 |
| | Accuracy | | | 99% | 1373 | Accuracy | | | 100% | 4788 | Accuracy | | | 99% | 1332 |
| | Macro Avg | 100% | 99% | 99% | 1373 | Macro Avg | 100% | 99% | 100% | 4788 | Macro Avg | 98% | 99% | 98% | 1332 |
| | Weighted Avg | 99% | 99% | 99% | 1373 | Weighted Avg | 100% | 100% | 100% | 4788 | Weighted Avg | 99% | 99% | 99% | 1332 |

The CNN-Integrated CCT model achieved consistently high performance across all five folds (see Table 3). For the 2-class. 3-class and 5-class breast cancer mammography accuracy remained between 99% and 100% in all folds. Slightly lower scores were observed for smaller classes, such as Mini-MIAS, where precision dropped to 93% in Fold 5, and INbreast, where recall ranged from 96% to 99%.

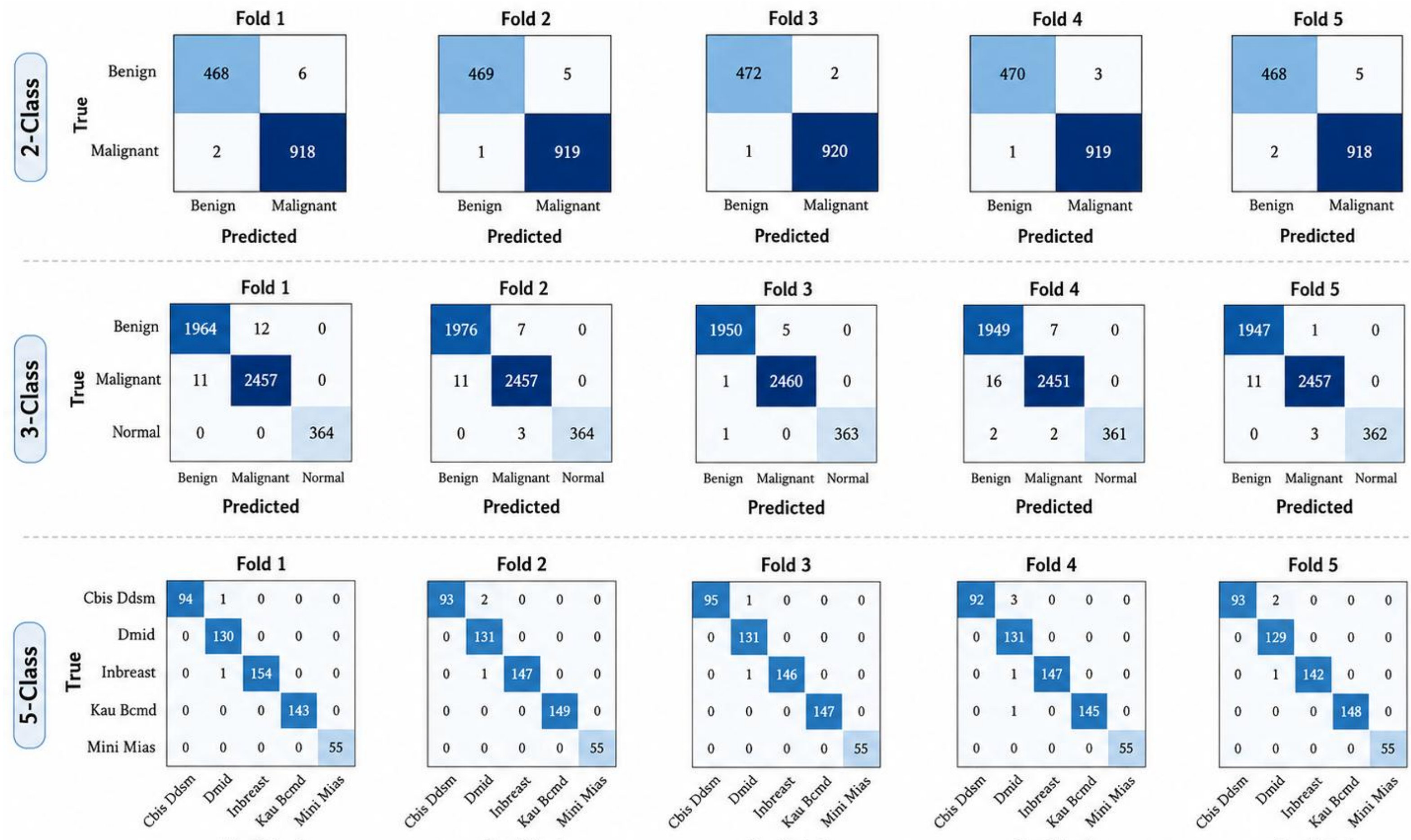

Figure 6: 5-Fold Confusion Matrix of 2-class. 3-class and 5-class Breast Mammography Classification

The confusion matrix in Figure 5 corresponds to the classification reports in Table 3. In the 2-class breast cancer classification, the model correctly classified around 468–472 benign and 918–920 malignant cases per fold, with only 3–8 total errors. This indicates the model effectively learned the features of benign and malignant cancer. In the 3-class breast cancer detection and classification task, the model correctly predicted about 2451–2460 malignant cases per fold, and the class ‘Normal’ was mostly correctly classified, around 361–364 samples. In the 5-class case, minor confusion occurs in smaller dataset groups such as CBIS-DDSM, INbreast, and KAUBCMD.

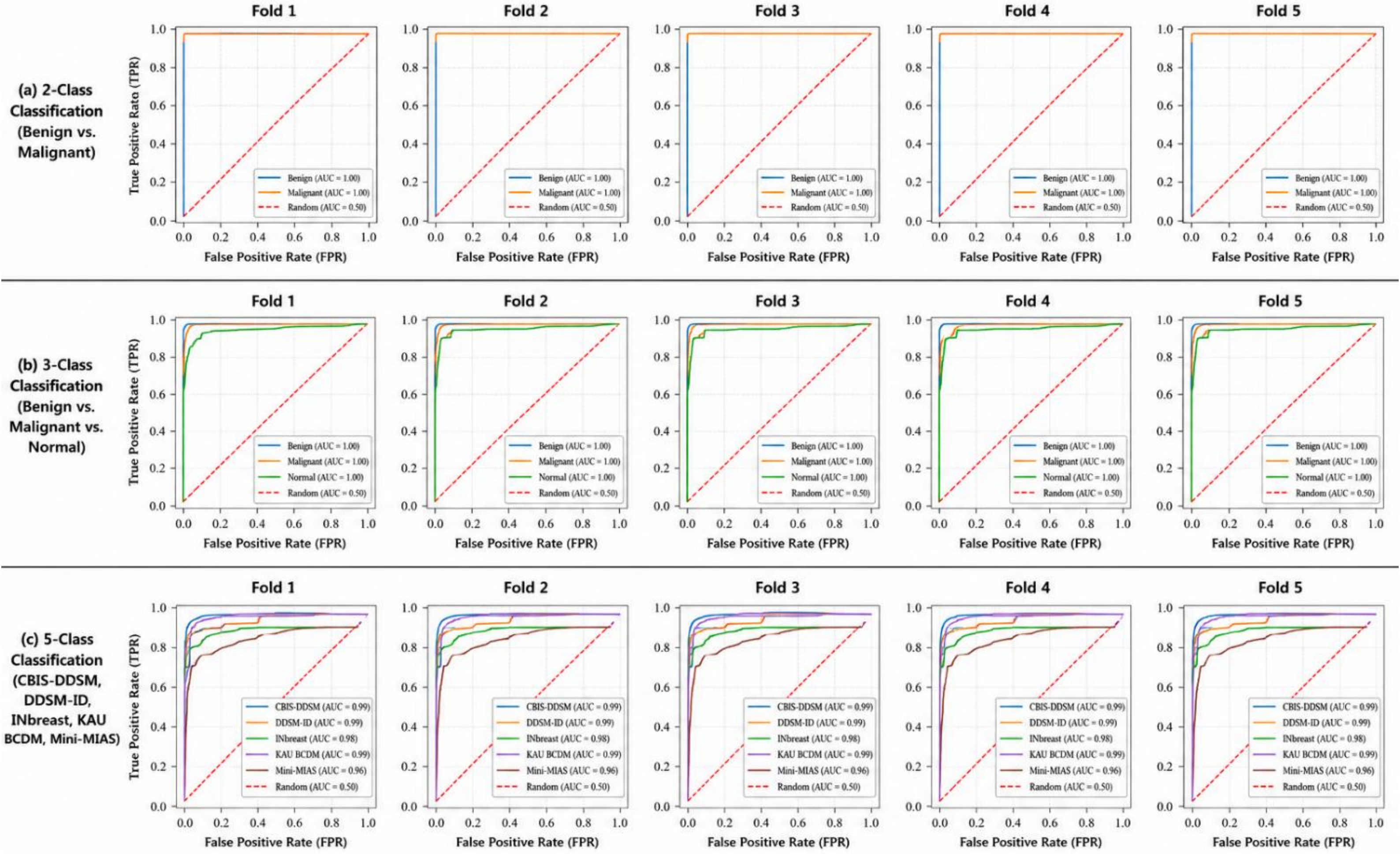

Figure 7: 5-Fold ROC-AUC Analysis of 2-class, 3-class, and 5-class Breast Mammography Classification

The ROC curves show strong and stable performance of the classification model across different classes (see Figure 7). In the 2-class breast cancer classification, both benign and malignant classes achieved an AUC of 1.00 across all folds, indicating excellent separation between cancerous and non-cancerous cases. The 3-class model also maintained AUC values of 1.00 for ‘Benign’, ‘Malignant’, and ‘Normal’ classes, showing that adding the normal category did not reduce classification performance. In the more challenging 5-class dataset-level classification, the model still performed strongly, with AUC values mostly between 0.98 and 0.99, while Mini-MIAS showed the lowest but still high AUC of about 0.96. Overall, the curves remain close to the upper-left corner and far above the random baseline of 0.50, supporting the robustness and generalization ability of the lightweight CNN embedding model.

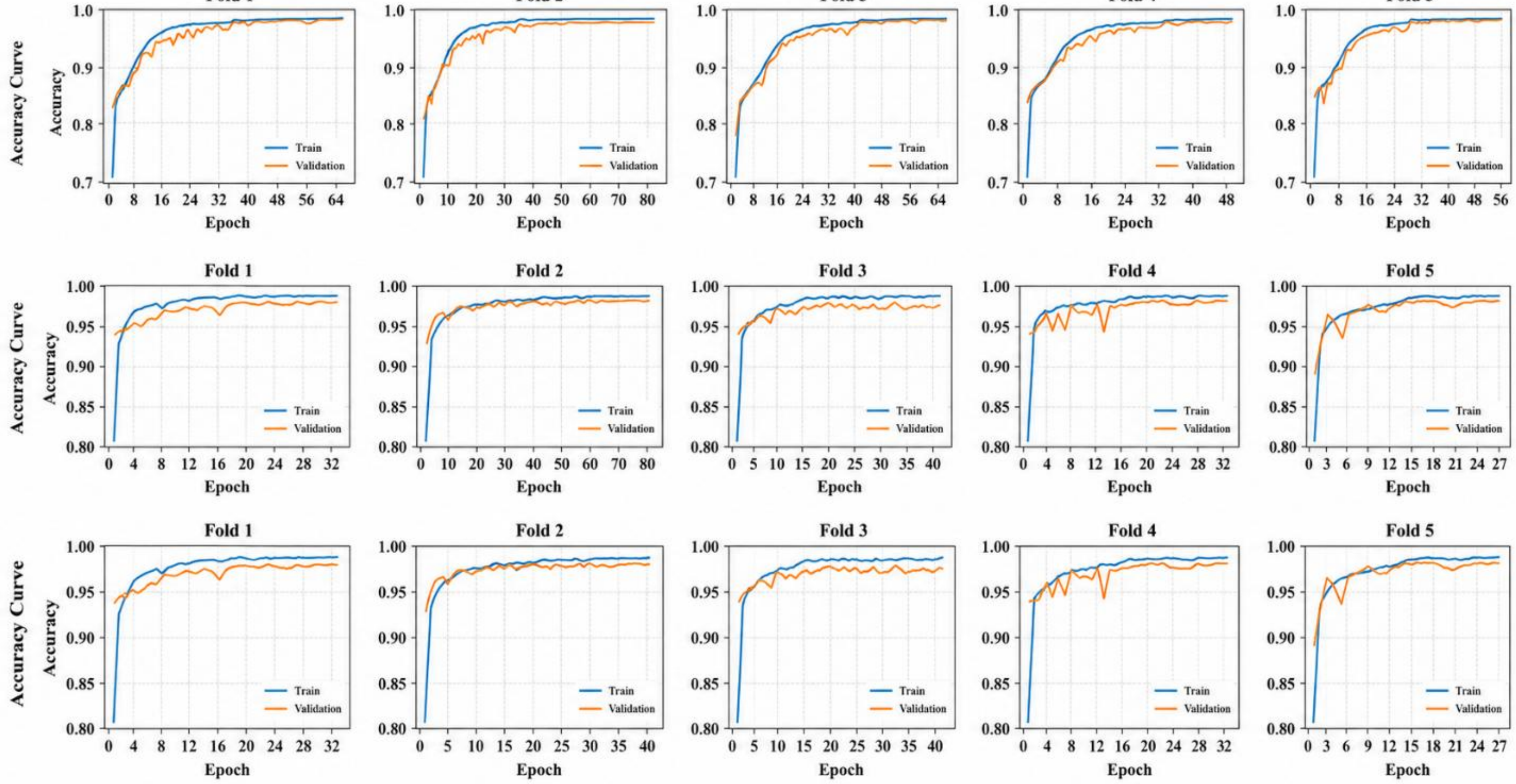

Figure 8: 5-Fold Training and Validation Accuracy Curves of CNN-Integrated CCT for Breast Mammography Classification

The training accuracy curve rises sharply from about 0.70–0.85 at the initial epochs and reaches nearly 0.98–1.00 in most folds (see Figure 8). Validation accuracy follows a similar pattern, stabilizing around 0.97–0.99, especially for Fold 4 and Fold 5. The narrow gap between training and validation curves suggests that the model learns effectively without strong overfitting.

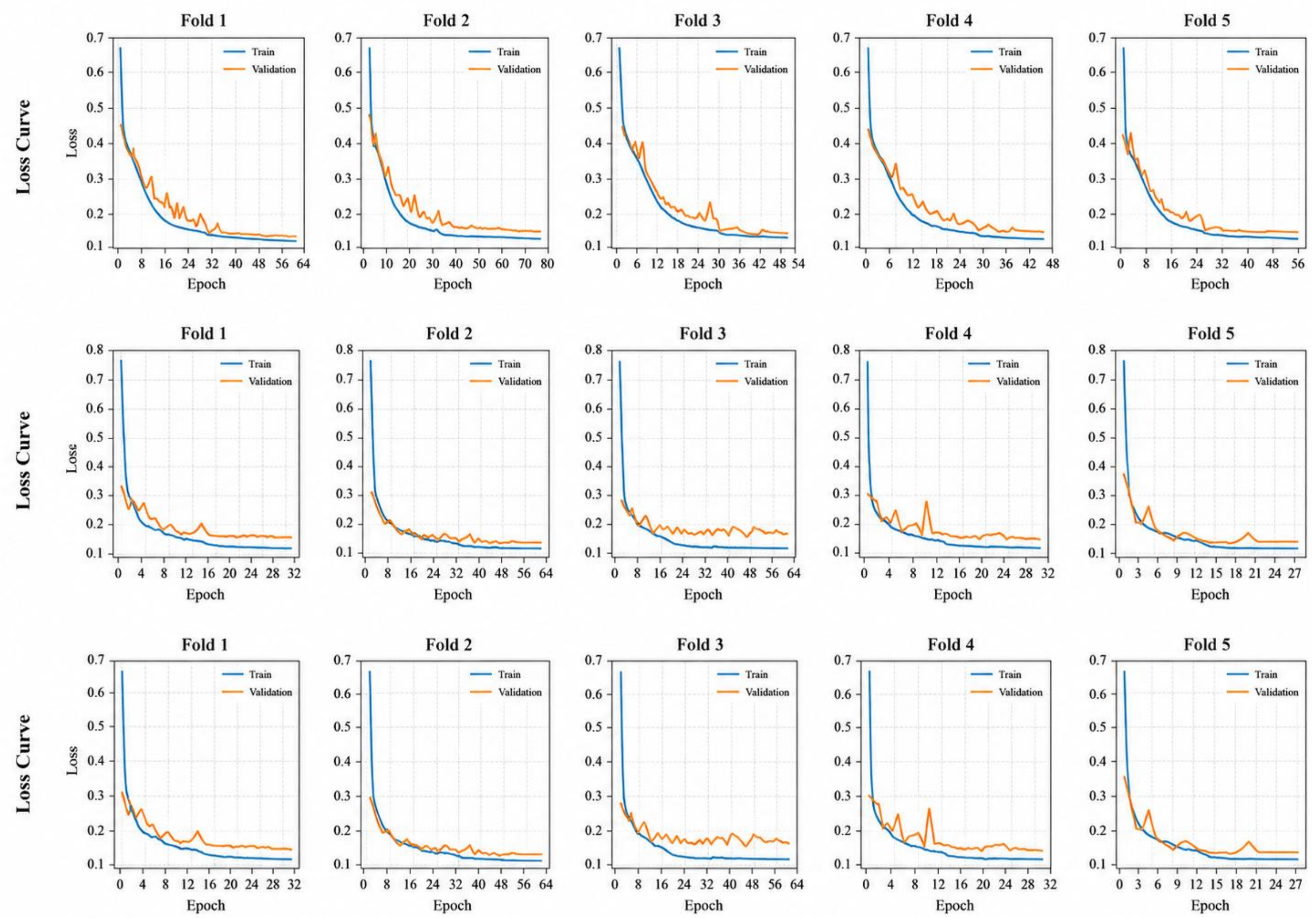

Figure 9: 5-Fold Training and Validation Loss Curves of CNN-Integrated CCT

Table 9 presents the training and validation loss. The loss curves show stable convergence across all five folds, with both training and validation losses decreasing rapidly in the early epochs. Initial loss values start around 0.65–0.80 for training and gradually decline to approximately 0.10–0.15, while validation loss stabilizes at 0.13–0.17. A few temporary fluctuations in validation are visible, particularly in Fold 3, Fold 4, and Fold 5, but they settle quickly without sustained divergence.

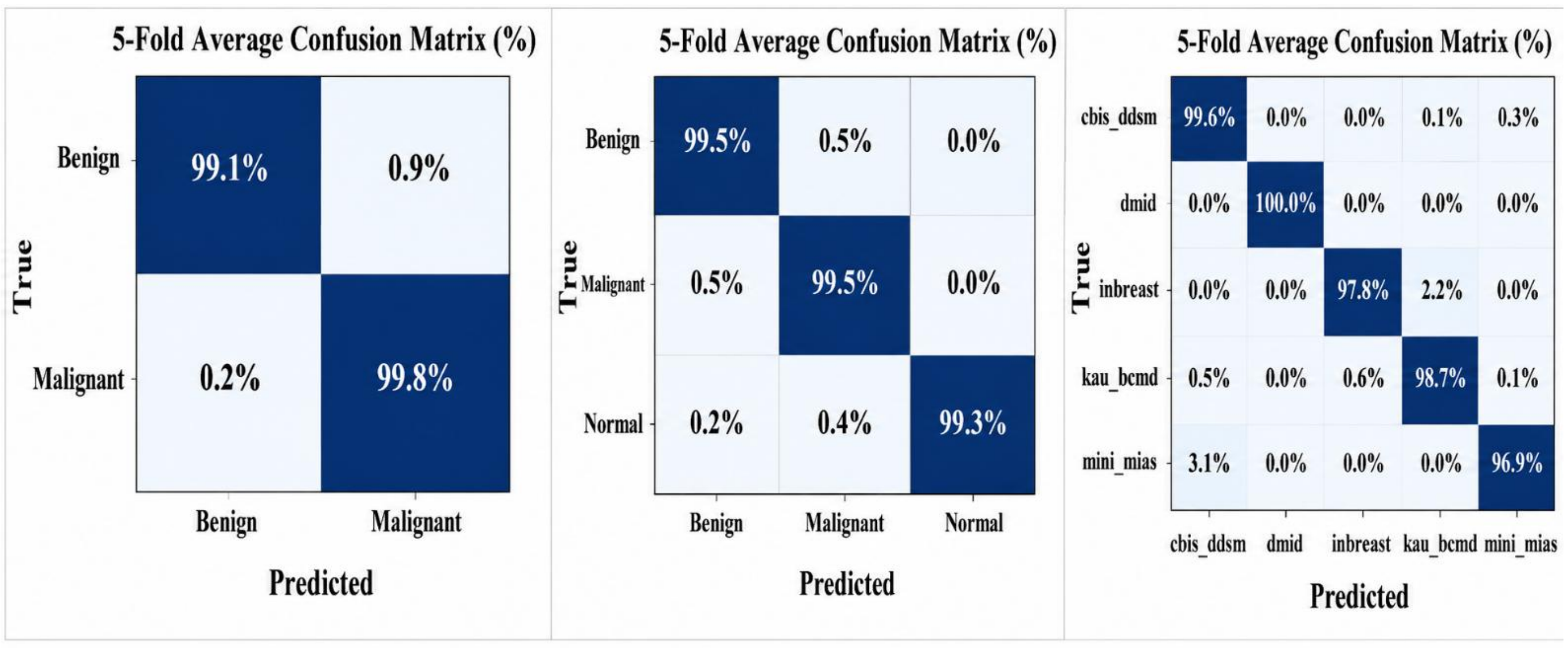

Figure 10: 5-Fold Average Confusion Matrices for Lightweight CNN-Based Breast Mammography Classification

The averaged confusion matrices are presented in Figure 10. In the 2-class classification, the model correctly identified 99.1% of the Benign class and 99.8% of the Malignant class. For the 3-class cancer detection and classification, the correct prediction rates of 99.5% for benign, 99.5% for malignant, and 99.3% for normal. The 5-class task was more challenging but still robust, with class-wise accuracies ranging from 96.9% to 100%; the lowest value was observed for Mini-MIAS, likely due to its smaller sample size.

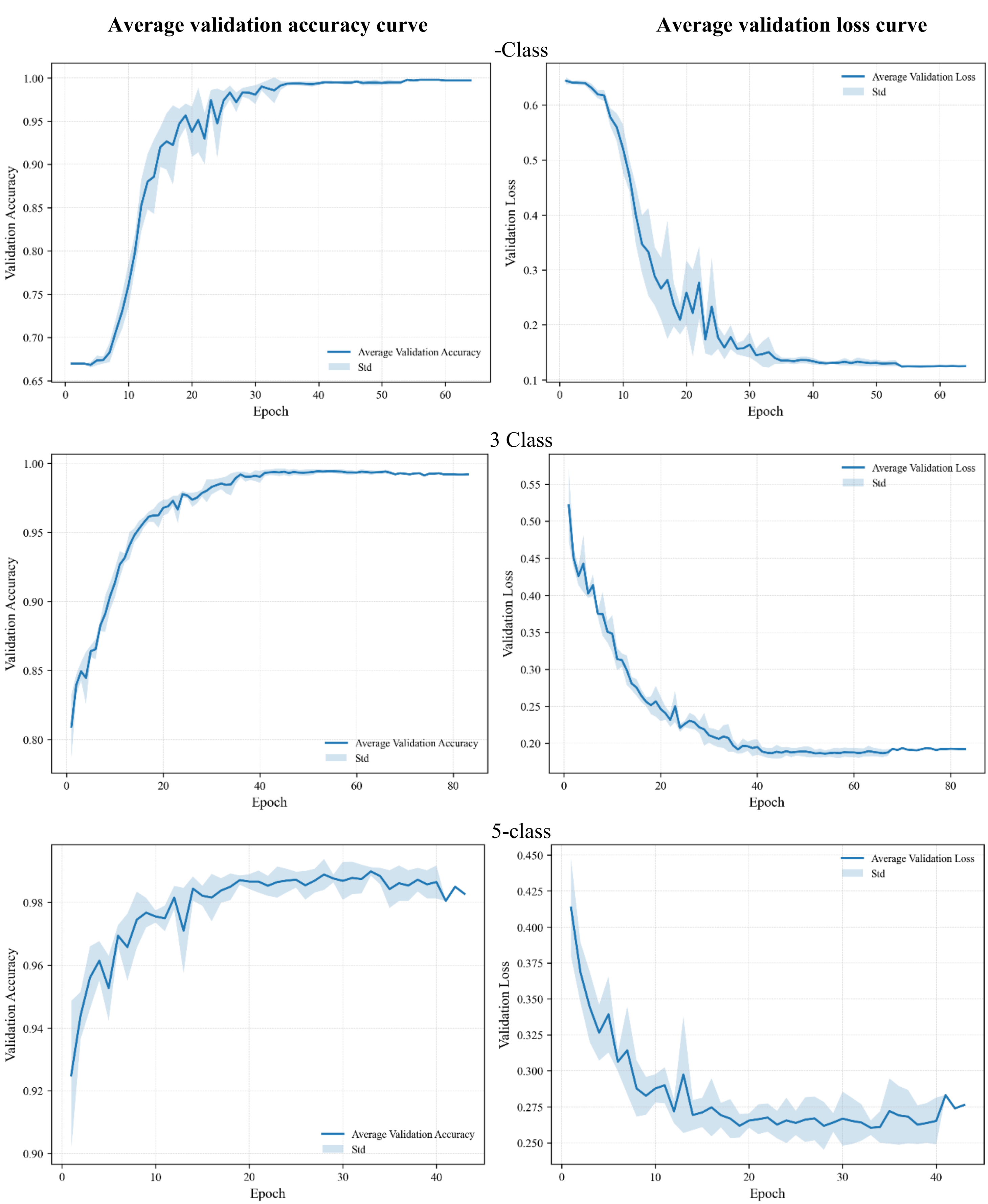

Figure 11: Average Validation Accuracy and Loss Curves for 2-Class, 3-Class, and 5-Class

The proposed CNN-integrated CCT model exhibits consistent performance for 2-, 3-, and 5-class breast cancer classification, with an average validation accuracy of 0.99 (±0.01) and a low, stable validation loss (see Figure 11). In the 5-class breast cancer classification, the model shows only a slight drop in accuracy due to increased class imbalance; accuracy remains above 0.98. The learning curves indicate rapid convergence in the early epochs and very stable behavior thereafter. The important aspect of the model is that there are no signs of overfitting, as the validation loss closely tracks the accuracy trends. Even in the case of 5-class classification, the model maintains high accuracy (>98%), indicating strong feature extraction capability and robustness.

## 4.2 Performance of unseen data detection and classification

Table 4: Performance of the Proposed Model on Unseen Test Data for Multi-Class Breast Cancer Classification

| **Classification Task** | **Test Accuracy (Mean ± SD)** | **Test Loss (Mean ± SD)** | **95% CI (Accuracy)** |
|---|---|---|---|
| 2-Class | 0.9955 ± 0.0024 | 0.1287 ± 0.0056 | [0.993, 0.998] |
| 3-Class | 0.9949 ± 0.0008 | 0.1844 ± 0.0030 | [0.994, 0.996] |
| 5-Class | 0.9870 ± 0.0016 | 0.2640 ± 0.0049 | [0.985, 0.989] |

The CNN-integrated CCT model performs very strongly on unseen test data, maintaining consistently high accuracy above 98% (see Table 4). The best performance is observed in the 2-class and 3-class tasks; a slight drop in performance appears in the 5-class case due to increased class complexity. The model's performance is further confirmed using a confusion matrix, ROC curve, and training and validation loss curves.

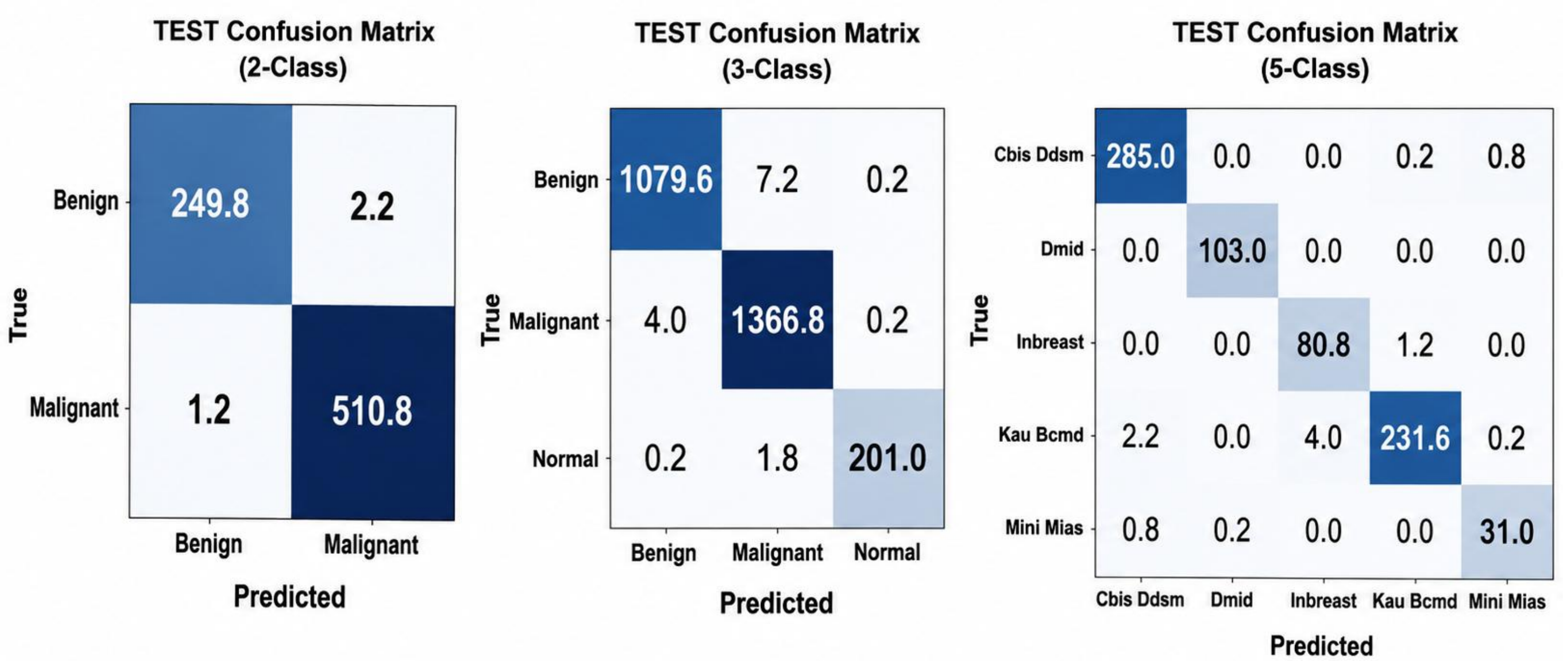

Figure 12: Comprehensive Confusion matrix of CNN-integrated CCT Performance (2-Class, 3-Class, and 5-Class Evaluation)

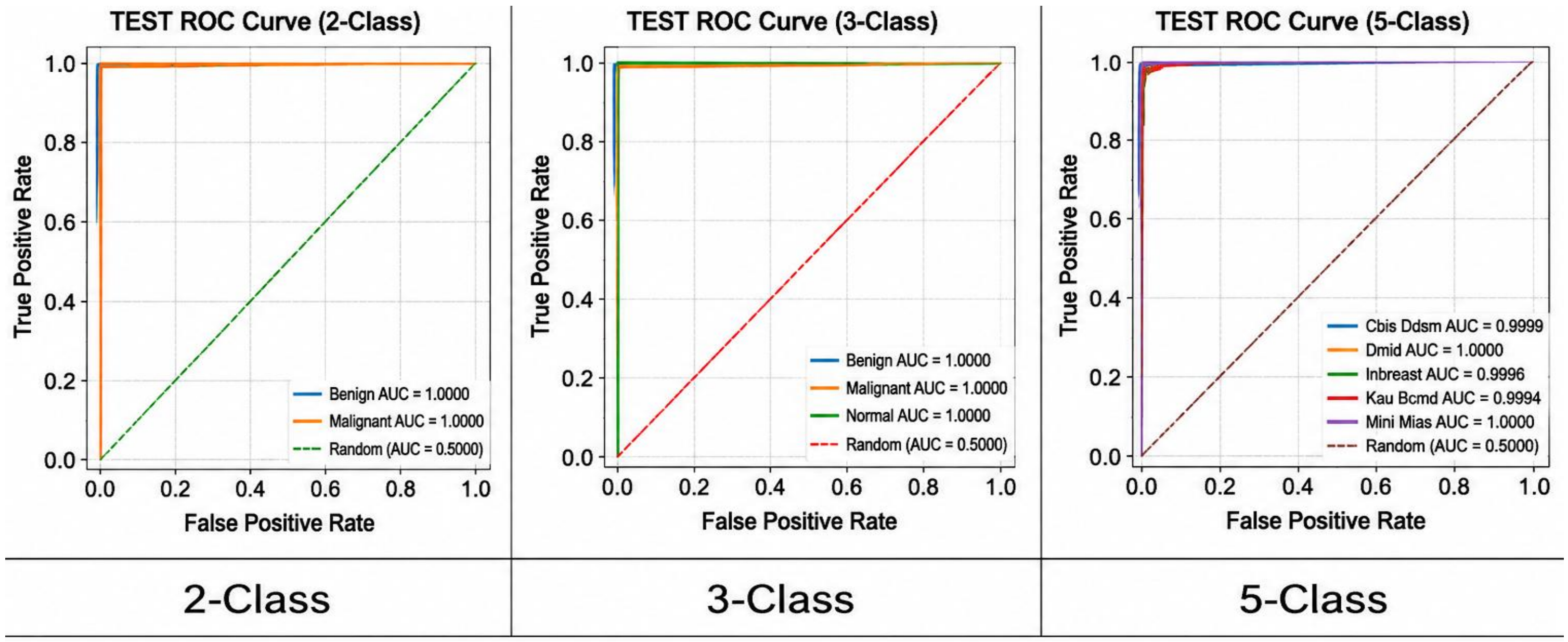

Figure 13: Comprehensive ROC Curve Analysis for Multi-Class Classification Performance (2-Class, 3-Class, and 5-Class Evaluation)

Figure 12 presents the confusion matrices for the test datasets of 2-class, 3-class, and 5-class breast cancer classification. The ROC curves also support the results from the Confusion Matrix. The ROC curve analysis suggests that in the 2-class and 3-class classification, the model achieved an AUC of 1.0000 (see Figure 12). For the 5-class classification task, the AUC values remain exceptionally high, ranging from 0.9994 to 1.0000, indicating near-perfect classification even with increased class complexity. Across all experiments, ROC curves remain tightly aligned toward the top-left corner of the plot, indicating very high true positive rates with minimal false positives.

2-class

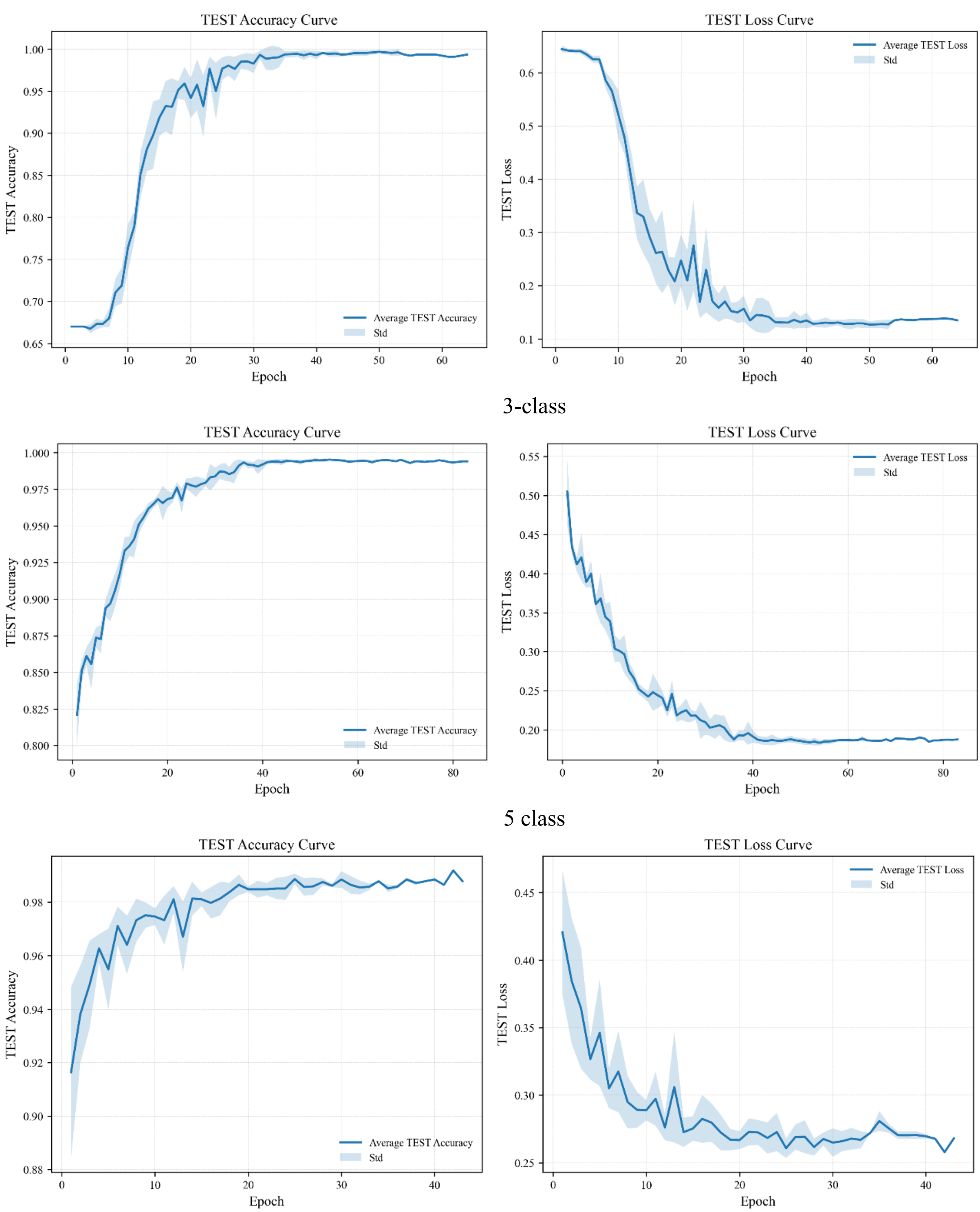

Figure 14: Comprehensive K-Fold Evaluation of Test Accuracy and Loss Across

Furthermore, the K-fold averaged test performance indicates that, in the 2-class model, test accuracy rises sharply in the initial epochs and reaches approximately 0.99 by around 25 epochs (see Figure 14). However, the test loss decreases steadily and stabilizes near 0.12, indicating fast learning and efficient optimization. For the 3-class scenario, the model shows slightly slower but consistent convergence, with accuracy stabilizing near 0.99 by approximately 30 epochs and test loss reducing to about 0.19, reflecting increased task complexity while still maintaining strong predictive performance. In the 5-class classification task, convergence remains stable, though comparatively gradual, with accuracy settling between 0.98 and 0.99 and test loss stabilizing around 0.27, indicating the expected impact of higher class complexity on model learning. Across all three experimental setups, the shaded variance bands remain narrow, confirming low inter-fold variability and robust generalization capability. Importantly, there is no clear evidence of overfitting, as both the accuracy and loss curves show synchronized, stable convergence.

## 4.3 Computational cost analysis of the CNN-integrated CCT Model

Table 5 summarizes the computational cost of the proposed CNN-integrated CCT model. The model contains only 0.2504 million parameters and requires less than 1 MB of storage, showing its lightweight design. The inference performance was also efficient. The model required approximately 1.4884 ms to process one image, with an increase in inference memory of only 0.9180 MB. The reported convolution-only computational cost was 0.00210944 GFLOPs, showing that the convolutional part of the model required very low computational effort. Since this value represents only the convolutional operations, it should be reported as conv-only GFLOPs rather than complete model GFLOPs.

Table 5. Computational efficiency profile of the proposed CNN-integrated CCT model

| Metric | Value | Interpretation |
|---|---|---|
| Total parameters | 250,435 | Indicates a lightweight architecture with few learnable weights. |
| Parameters | 0.2504 M | The model contains only about one-quarter million parameters, making it compact. |
| Model size | 0.9553 MB | The trained model requires less than 1 MB of storage, supporting deployment in resource-limited settings. |

| Inference latency | 1.4884 ms/image | The model produces predictions rapidly for each input image. |
|---|---|---|
| FLOPs | 0.00210944 G | Indicates low computational cost for the convolutional component. |
| GFLOPs | 0.00210944 | Confirms that the model requires very limited floating-point operations. |
| Inference memory delta | 0.9180 MB | Shows low additional memory use during inference. |

## 4.4 Statistical analysis of the result of 3-class breast mammography

The descriptive training statistics suggest that the mean training accuracy was 0.9488 and the median was 0.9905 (see Table 6). This indicates that the model achieved high accuracy during training. The mean training loss was 0.2628, while the median loss was 0.1952, showing that the loss gradually decreased as the model learned more discriminative features.

Table 6. Descriptive statistics of training accuracy and loss

| **Statistical Measure** | **Training Accuracy** | **Training Loss** |
|---|---|---|
| Mean | 0.9488 | 0.2628 |
| Standard deviation | 0.0691 | 0.1120 |
| Median | 0.9905 | 0.1952 |
| Interquartile range | 0.0857 | 0.1576 |
| Range | 0.4207 | 0.6277 |
| 95% confidence interval | 0.9348–0.9628 | 0.2401–0.2855 |

The normality tests showed that the epoch-wise accuracy and loss distributions were not normally distributed; therefore, the Wilcoxon test provided additional non-parametric support for the statistical consistency of the training results.

Table 7. Statistical testing summary of training behavior

| **Test / Measure** | **Training Accuracy** | **Training Loss** | **Interpretation** |
|---|---|---|---|
| t-statistic | 134.5178 | 22.9966 | Both metrics showed statistically strong deviation from the tested reference level. |
| p-value | $3.86 \times 10^{-111}$ | $7.89 \times 10^{-41}$ | Very small p-values indicate statistically significant training trends. |
| Effect size | 13.7292 | 2.3471 | Accuracy showed a very large effect size, and loss showed a strong effect as well. |
| Bias-corrected effect size | 13.6205 | 2.3285 | Confirms the strength of the observed training behavior after correction. |
| Normality statistic | 0.7322 | 0.7718 | Values suggest that the epoch-wise distributions were not normally distributed. |
| Normality p-value | $5.20 \times 10^{-12}$ | $5.75 \times 10^{-11}$ | The low p-values indicate significant deviation from normality. |
| Wilcoxon statistic | 0.0000 | 0.0000 | Non-parametric testing supports the statistical consistency of the training trend. |
| Wilcoxon p-value | $1.22 \times 10^{-17}$ | $1.22 \times 10^{-17}$ | The non-parametric results were statistically significant. |

Training a 3-class breast mammography model suggests a strong negative relationship between accuracy and loss. The p-values of the Pearson correlation coefficient (-0.9966) and the Spearman correlation coefficient (-0.9981) confirm that, during training, accuracy increased and loss decreased. This suggests a consistent, expected learning pattern for CNN-integrated CCT from breast mammography.

Table 8. Relationship between training accuracy and loss

| **Correlation Type** | **Correlation Coefficient** | **p-value** | **Interpretation** |
|---|---|---|---|
| Pearson correlation | -0.9966 | $5.30 \times 10^{-105}$ | Shows a very strong inverse linear relationship between accuracy and loss. |
| Spearman correlation | -0.9981 | $5.70 \times 10^{-117}$ | Confirms a very strong monotonic inverse relationship. |

Table 9. Ablation Study of the Proposed CNN–integrated CCT Model

| Experiment | Train Acc. | Eval Acc. | Test Acc. | Train Loss | Eval Loss | Test Loss | Macro AUC | Best Eval Acc. | Best Eval Loss | Epochs | Aug | CNN Stem | Pos. Emb. | Stoch. Depth | Layers | Dim | Heads | Label Smooth |
|---|---|---|---|---|---|---|---|---|---|---|---|---|---|---|---|---|---|---|
| Baseline CNN–CCT | 99.98% | 98.78% | 98.65% | 0.225 | 0.257 | 0.261 | 0.9989 | 98.99% | 0.257 | 47 | ✓ | ✓ | ✓ | ✓ | 2 | 64 | 4 | 0.05 |
| No Stochastic Depth | 99.98% | 99.32% | 99.05% | 0.224 | 0.245 | 0.261 | 0.9999 | 99.32% | 0.245 | 67 | ✓ | ✓ | ✓ | ✕ | 2 | 64 | 4 | 0.05 |
| Shallow Transformer (1 Layer) | 99.98% | 99.26% | 98.92% | 0.226 | 0.244 | 0.253 | 0.9999 | 99.26% | 0.244 | 42 | ✓ | ✓ | ✓ | ✓ | 1 | 64 | 4 | 0.05 |
| Fewer Attention Heads (2) | 99.96% | 99.12% | 98.79% | 0.226 | 0.253 | 0.255 | 0.9999 | 99.12% | 0.253 | 38 | ✓ | ✓ | ✓ | ✓ | 2 | 64 | 2 | 0.05 |
| No Positional Embedding | 99.98% | 98.78% | 98.65% | 0.225 | 0.250 | 0.254 | 0.9997 | 98.99% | 0.250 | 63 | ✓ | ✓ | ✕ | ✓ | 2 | 64 | 4 | 0.05 |
| No Label Smoothing | 99.98% | 98.71% | 98.65% | 0.0013 | 0.054 | 0.042 | 0.9999 | 98.78% | 0.054 | 32 | ✓ | ✓ | ✓ | ✓ | 2 | 64 | 4 | 0 |
| No CNN Stem | 99.03% | 98.51% | 98.38% | 0.251 | 0.261 | 0.262 | 0.9996 | 98.58% | 0.261 | 77 | ✓ | ✕ | ✓ | ✓ | 2 | 64 | 4 | 0.05 |
| No Data Augmentation | 99.98% | 98.78% | 98.25% | 0.226 | 0.264 | 0.274 | 0.9995 | 98.91% | 0.264 | 26 | ✕ | ✓ | ✓ | ✓ | 2 | 64 | 4 | 0.05 |

To systematically evaluate the contribution of each architectural component in the proposed CNN–CCT framework, an extensive ablation study was conducted, as summarized in Table 9. The results demonstrate that the full model achieves the most balanced performance across accuracy, loss stability, and Macro AUC, confirming the effectiveness of the hybrid CNN–Transformer design.

The removal of stochastic depth leads to a slight improvement in evaluation accuracy; however, this is accompanied by less stable convergence behavior across epochs, indicating reduced regularization effectiveness. Similarly, reducing the transformer depth to a single layer maintains competitive accuracy but slightly limits the model's ability to capture deep global contextual dependencies. This confirms that even shallow transformer structures can perform well when supported by strong CNN feature extraction, but additional depth enhances representational richness.

Reducing the number of attention heads results in a minor decline in test performance, highlighting the importance of multi-head self-attention in capturing diverse feature subspaces within mammographic representations. In contrast, removing the CNN stem produces a more noticeable drop in performance, confirming that convolutional inductive bias is essential for extracting local texture patterns such as lesion boundaries and intensity variations.

The absence of positional embeddings also negatively impacts performance, demonstrating that spatial encoding remains critical even in compact token-based architectures. Furthermore, removing data augmentation leads to the most significant degradation in generalization, as reflected in increased test loss and reduced accuracy, underscoring its role in mitigating dataset bias and improving robustness.

Finally, disabling label smoothing yields highly confident but less well-calibrated predictions, as evidenced by extremely low training loss and reduced generalization performance. Overall, the ablation study confirms that each component contributes synergistically to the final performance and that integrating CNN feature extraction with transformer-based global reasoning yields a robust and clinically reliable classification framework.

## XAI integration analysis

Malignant: 97% Malignant: 97% Normal: 97.2% Benign: 97.6%

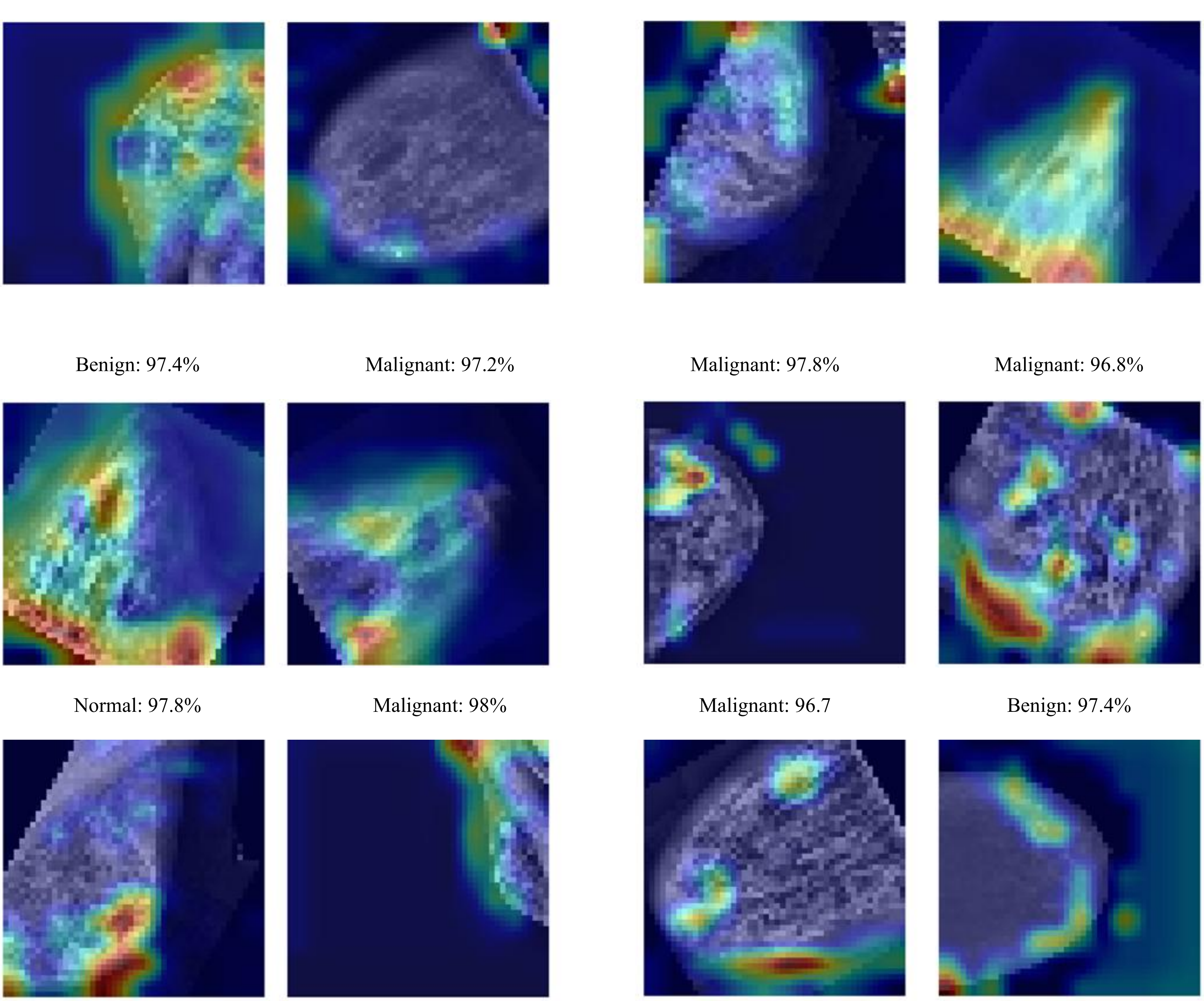

Figure 15. Grad-CAM visualizations with class probability scores

The Grad-CAM visualizations provide qualitative evidence of the model's decision-making process (see Figure 15). The class probability scores demonstrate that the proposed CNN-integrated CCT model produced confident predictions for Benign, Malignant, and Normal samples. The highlighted activation regions were mostly located within tissue-containing regions, suggesting that the model relied on relevant visual patterns rather than background information. The Grad-CAM heatmaps show that the model focused mainly on visible tissue regions rather than empty background areas. For Malignant cases, the highlighted regions were generally concentrated in dense, irregular tissue patterns. For Benign and Normal cases, the activation regions appeared more localized and less aggressive in distribution. These visual results suggest that the model learned meaningful image features and relied on diagnostically relevant regions for its predictions.

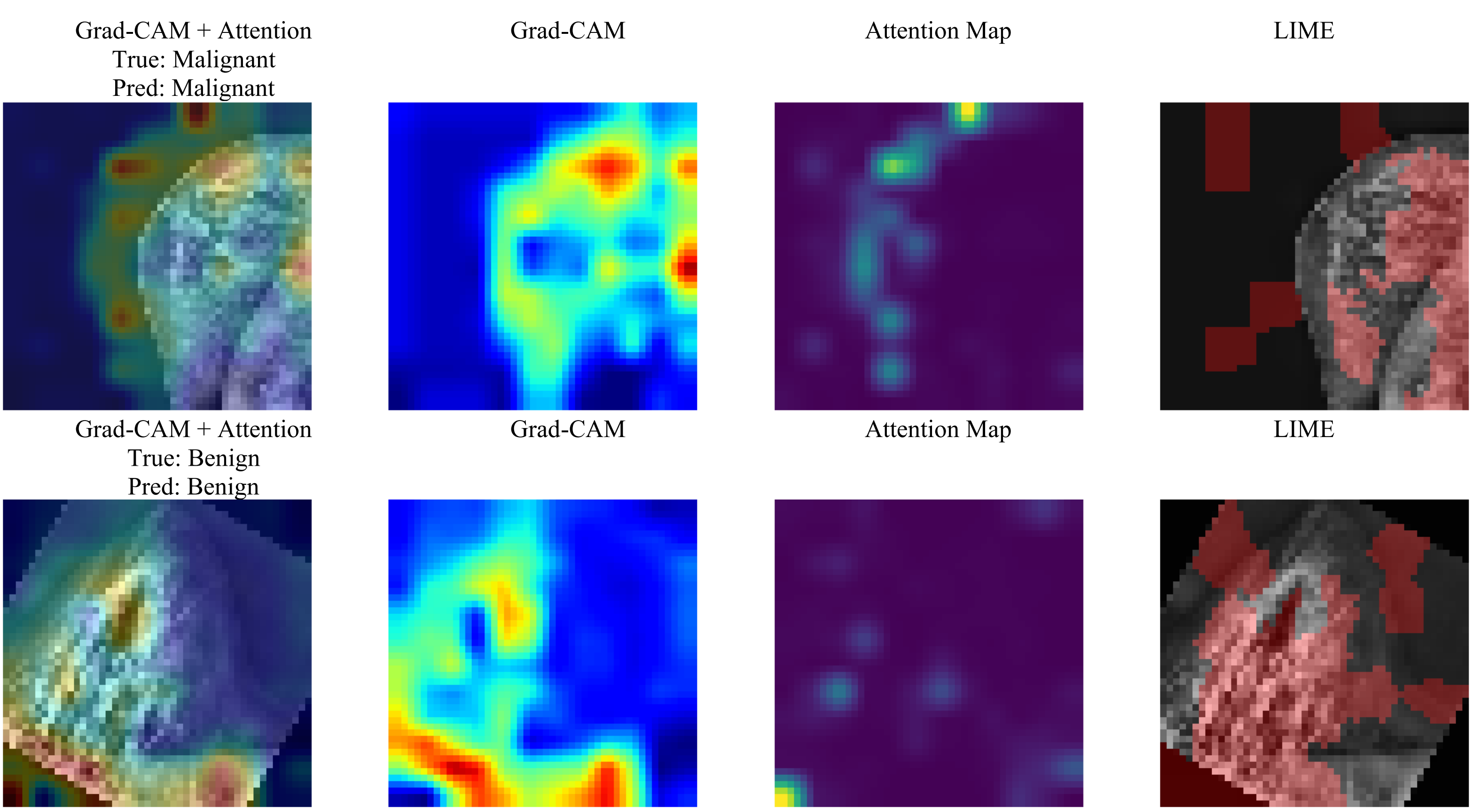

Figure 16: Comparative explainability visualization of the proposed CNN-integrated CCT model using fused Grad-CAM–attention maps, Grad-CAM, attention maps, and LIME

Figure 16 shows the explainability results of the proposed CNN-integrated CCT model for representative breast image samples. The visualization results indicate that the model mainly focused on informative tissue regions rather than background areas. For example, for a Malignant sample, the highlighted regions were mostly located around dense and irregular tissue patterns. The highlighted regions indicate image areas that contributed most strongly to the model prediction. The results show that the proposed model focused on relevant breast tissue regions, supporting the transparency and reliability of the classification decision.

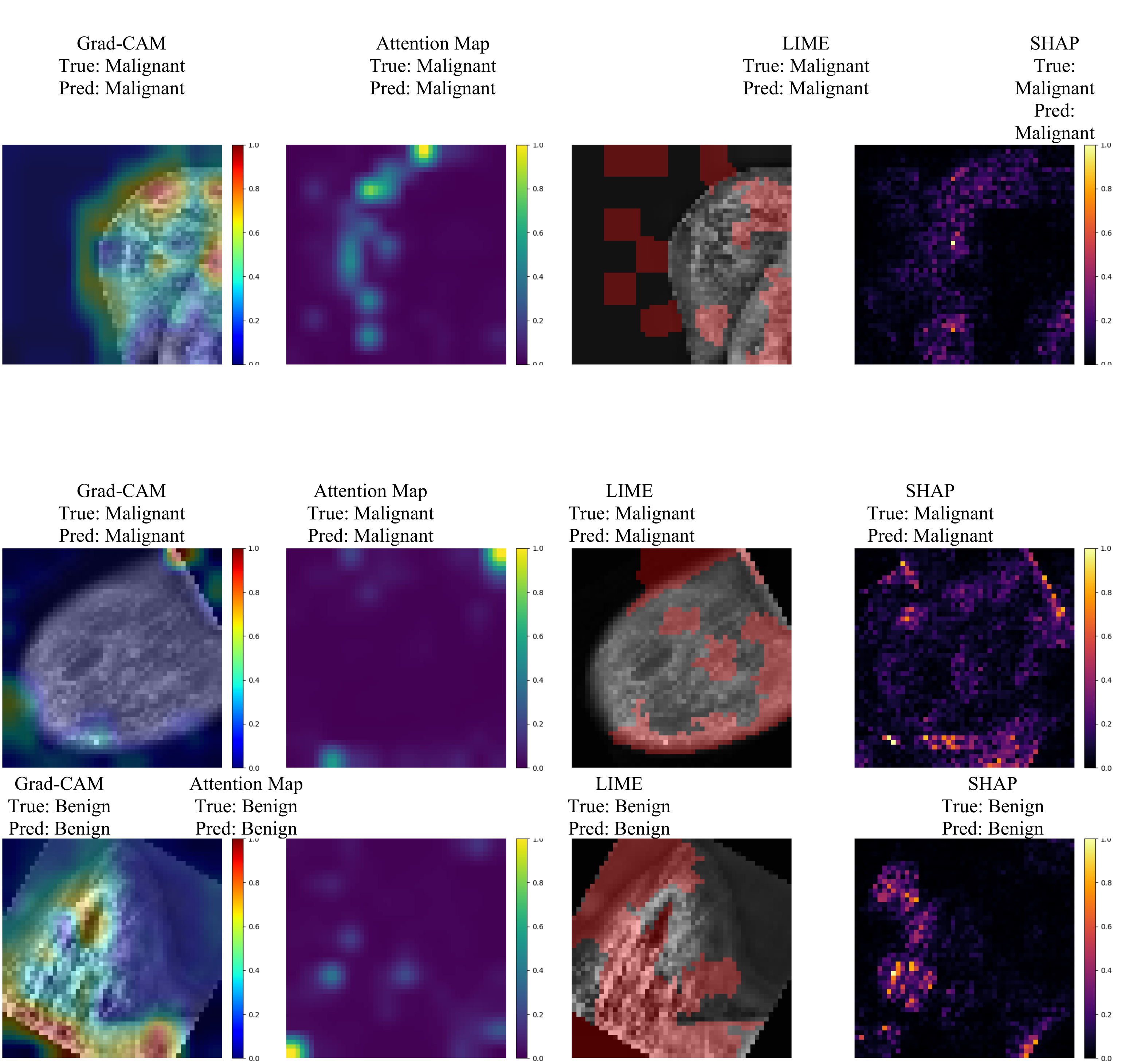

Figure 17: Explainable artificial intelligence visualization of the proposed model

Figure 17 combines Grad-CAM, attention map, LIME, and SHAP explanations. The color scale ranges from low to high contribution, with brighter regions indicating a stronger influence on the model prediction. The visual explanations show that the proposed model focused on relevant breast tissue regions for Benign and Malignant classification, supporting the interpretability and trustworthiness of the prediction process.

# 5. Discussion

Significant studies applied DL methods to detect and classify breast cancer. Our proposed CNN-integrated CCT consistently achieves strong performance across 5-fold cross-validation experiments on 2-class, 3-class, and 5-class mammographic datasets. The model achieved 99%–100% accuracy across 3 datasets, indicating robust generalization. Moreover, performance stability across folds suggests that the hybrid architecture effectively mitigates overfitting despite dataset imbalance and inter-dataset variability. However, slight performance degradation was observed on smaller, more challenging datasets such as Mini-MIAS and INbreast, where recall values dropped to 93–97% across certain folds. This indicates that class scarcity and domain-specific imaging variability continue to influence feature separability, particularly under low-sample conditions.

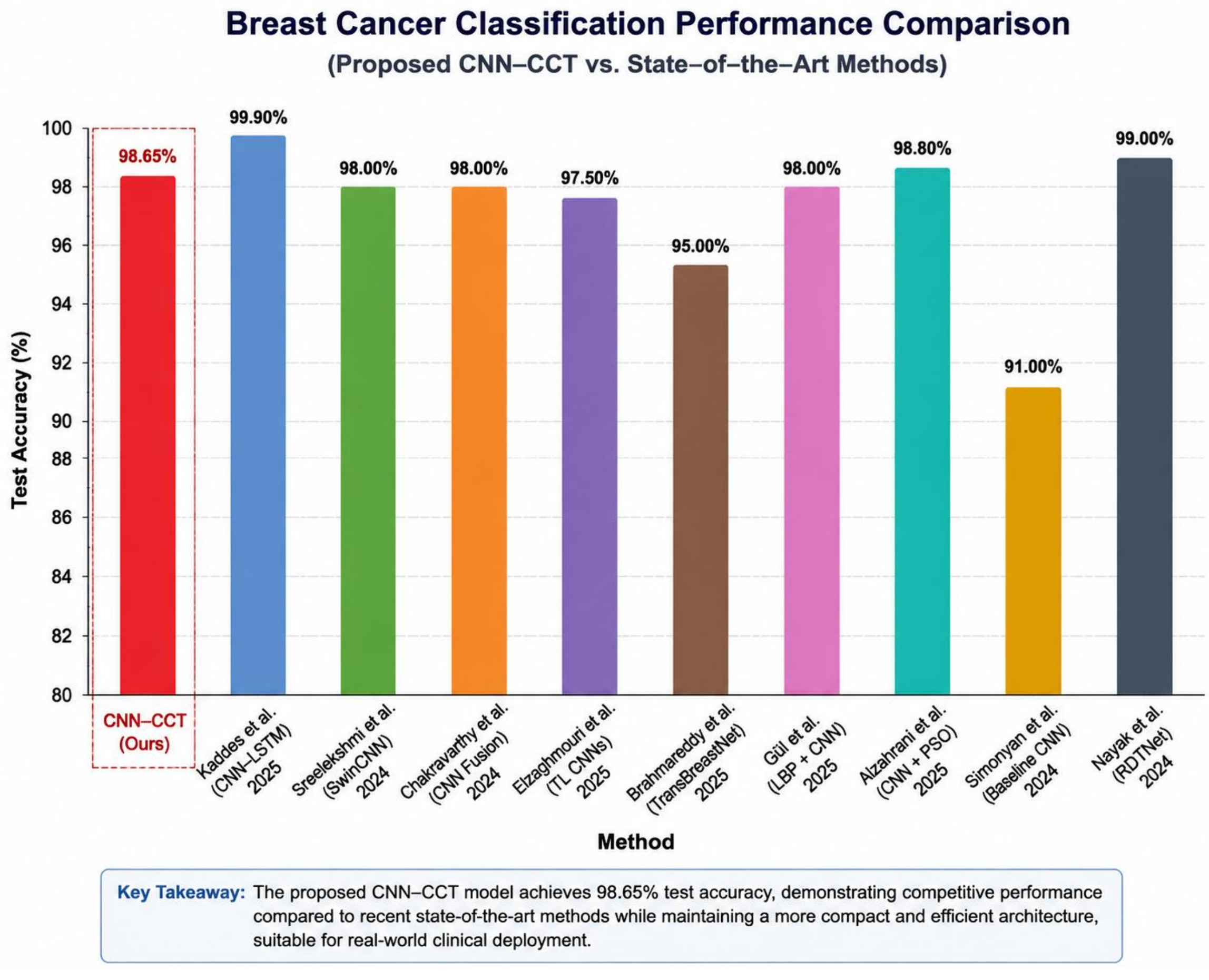

Figure 18: Breast Cancer Classification Performance Comparison

Comparing the recent state-of-the-art studies, our model demonstrates competitive or superior performance. For instance, CNN–LSTM architectures (Kaddes et al., 2025) achieve extremely high accuracy (~99.9%) but are limited by modality constraints and lack generalization across diverse datasets. Similarly, Swin CNN-based hybrid transformers (Sreeleklekshmi et al., 2024) and ViT-CNN architectures (Abimouloud et al., 2024) achieve strong accuracy (~98%), but at the cost of high computational complexity and increased training overhead. In contrast, the proposed CNN–integrated CCT model achieves comparable or improved accuracy while maintaining a more compact and efficient transformer design.

Traditional CNN-based approaches, such as Simonyan et al. (2024) and ensemble fusion models (Chakravarthy et al., 2024), generally achieve lower or comparable accuracy (90–98%), primarily due to limited global contextual modeling. Likewise, wavelet-CNN and PSO-optimized CNN models (Bohra et al., 2026; Alzahrani et al., 2025) improve feature extraction but incur additional computational cost and do not fully address long-range dependency modeling.

The superior performance of the proposed framework can be attributed to three key design advantages. First, the CNN backbone efficiently captures fine-grained spatial texture patterns such as microcalcifications and mass boundaries. Second, the CCT tokenizer reduces computational redundancy by converting feature maps into compact token representations. Third, the transformer encoder enhances global contextual reasoning through multi-head self-attention, enabling better discrimination between visually similar classes.

Despite these strengths, certain limitations persist. Performance variability in smaller datasets indicates that the model still depends on sufficient training diversity. Additionally, although transformer layers are computationally more efficient than standard Vision Transformers, their inclusion still introduces non-trivial inference overhead compared to pure CNN models.

Overall, the results confirm that integrating convolutional inductive bias with transformer-based global representation learning yields a balanced and effective solution for medical image classification. The CNN–integrated CCT framework demonstrates strong potential for clinical decision support systems, particularly in breast cancer screening applications where both accuracy and robustness are critical.

Prior studies emphasized the need for DL on edge devices for real-time detection and monitoring of breast cancers. Figure 18 compares the parameter counts of SOTA CNNs and CNN-integrated CCT, demonstrating that CNN-integrated CCT is a suitable model for edge devices. This is because the model will require fewer computational resources than the CNN. The XAI implementation is not limited to generating heatmaps using LIME, SHAP, and Grad-CAM. Instead, our XAI implementation is much more extensive than only generating heatmaps.

# 6. Limitations, Future, and Conclusions

Despite the promising performance of the proposed CNN-integrated CCT model for breast mammography image classification, several limitations should be acknowledged. The main limitation of this study is that it utilized a secondary dataset to evaluate the performance of the CNN-integrated CCT model in breast cancer detection and classification. However, we would like to point out that primary data collection is difficult in the area where this research was conducted. Second, this research considered only mammographic images of breast cancer; however, the model can be tested on Magnetic Resonance Imaging (MRI), Computed Tomography (CT), Ultrasound (US), X-ray Imaging (XR), Plasmid Bluescript (PBS), and Microscopy images. Future studies, therefore, should conduct extensive multi-center validation using the model across heterogeneous datasets.

There is room for improvement in the model as well. Since the model training process starts with input augmentation of mammography images, the quality and diversity of the input mammographic dataset are important. Any bias in the input image may propagate through subsequent CCT tokenizer and transformer encoder stages, potentially limiting generalization to unseen clinical trial settings. In short, variations in mammographic devices, acquisition protocols, and operator-dependent artifacts may reduce robustness when deployed in real-world hospital environments. Although the CNN backbone reduces the spatial complexity of the input mammography image before tokenization, the transformer encoder may introduce additional computational overhead due to multi-head self-attention and stochastic depth regularization. This increases inference latency compared to purely convolutional architectures, potentially restricting deployment in resource-constrained or real-time diagnostic systems. Future studies should optimize the transformer encoder by using lightweight attention mechanisms, such as linear attention, sparse attention, or token pruning. The CCT tokenizer and positional embedding module assume a fixed patch-based representation of CNN feature maps. This fixed tokenization strategy may result in suboptimal representation of fine-grained tumor boundaries, particularly when lesion morphology is highly irregular or low-contrast. The model may be sensitive to hyperparameter configurations, including the embedding dimension (D), the number of transformer layers (L), the patch size, and the dropout rate. Suboptimal tuning may lead to instability in convergence or degraded classification performance.

The proposed CNN-integrated CCT framework effectively combines local mammographic feature extraction with global contextual learning, enabling the detection of subtle abnormalities while maintaining efficient transformer-based representation learning. The study demonstrates the potential of convolutional tokenization and compact transformer design for developing accurate and robust computer-aided breast cancer classification systems.

In the model. The CNN layer captures hierarchical local features, while the CCT tokenizer transforms these feature maps into compact patch embeddings, which are critical for accurate breast lesion characterization in mammographic imaging. Positional embeddings are used to preserve spatial information, and a lightweight transformer encoder is employed to model long-range contextual dependencies via multi-head self-attention. Finally, a classification head produces the diagnostic prediction for benign and malignant cases.

Experimental results indicate that the hybrid CNN-integrated CCT model achieves strong classification performance for 2-, 3-, and 5-class breast lesions. With only 250,435 parameters, the model is a suitable model for real-time clinical deployment, further enhancing its applicability in computer-aided diagnosis systems. The model was integrated with LIME, SHAP, and Grad-CAM to deepen our understanding of how it detects and classifies input images.

Finally, since many scholars advocate integrating AI for practicality, we present a prototype that combines the CNN-integrated CCT model, a Streamlit-based web interface, and an Android mobile application. The prototype is expected to support ongoing efforts in the biomedical field to make CAD more practical, interpretable, and widely usable. Furthermore, while the core focus is on breast cancer classification, the broader framework has potential applications in other areas of disease detection, including agricultural disease management, where similar challenges around accessibility and decision support persist.